\documentclass{article}

\usepackage[final]{corl_2026} 

\usepackage{booktabs}
\usepackage{multirow}
\usepackage{graphicx}
\usepackage{amsmath}
\usepackage{xcolor}
\usepackage{url}
\usepackage{subcaption}
\usepackage{graphicx}
\usepackage[table]{xcolor}
\usepackage{float}
\usepackage{wrapfig}
\usepackage{tabularx}
\usepackage{enumitem}
\usepackage{hyperref}
\usepackage{pdfpages}

\definecolor{skyblue1}{RGB}{235,245,255}
\definecolor{skyblue2}{RGB}{215,232,248}

\usepackage{pifont}
\newcommand{\cmark}{\textcolor{green!60!black}{\ding{51}}}  
\newcommand{\xmark}{\textcolor{red!70!black}{\ding{55}}}    

\usepackage{authblk}

\title{H2RBench: A Real-to-Sim Benchmark for Evaluating Human-to-Robot Transfer}

\author[1]{Chuyang Xiao\textsuperscript{*}}
\author[1]{Haotian Zhan\textsuperscript{*}}
\author[1]{Sriram Krishna}
\author[2]{Peilin Meng}
\author[3]{Muhammad Zubair Irshad}
\author[3]{Sergey Zakharov}
\author[1]{David Held}

\affil[1]{Robotics Institute, Carnegie Mellon University}
\affil[2]{University of Michigan}
\affil[3]{Toyota Research Institute}

\affil[ ]{\textsuperscript{*}Equal contribution; authors are listed in alphabetical order.}

\date{}

\begin{document}
\maketitle

\begin{figure*}[ht!]
\centering
\includegraphics[width=\linewidth]{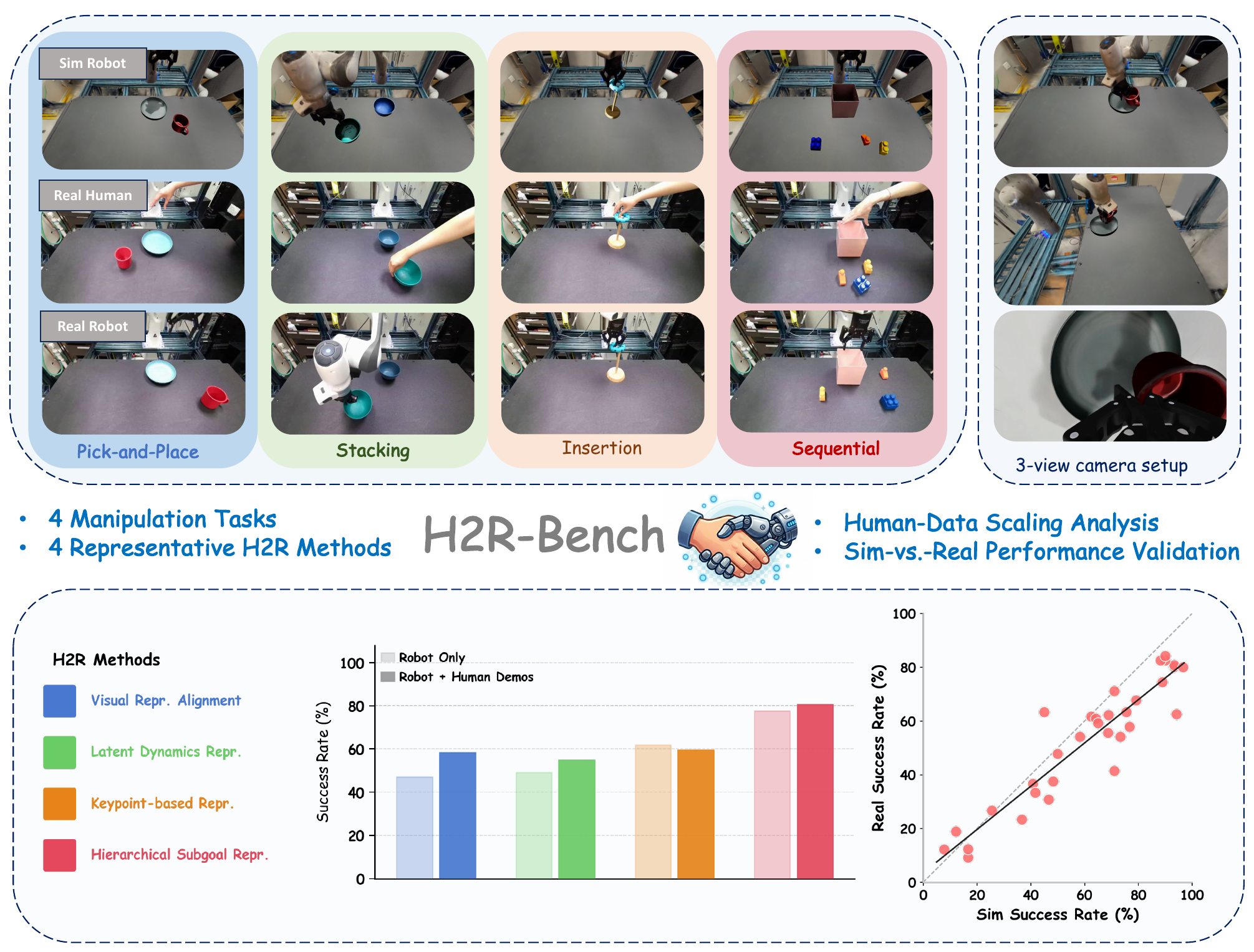}
\caption{\textbf{H2RBench overview.}
We propose a real-to-sim benchmark for evaluating human-to-robot
(H2R) policy transfer from real human video demonstrations.
H2RBench benchmarks multiple H2R methods across four
manipulation tasks, enabling systematic analysis of human-data
scaling. More videos are available on our \href{https://davidz2851.github.io/h2rbench.github.io/}{project website}.}
\label{fig:teaser}
\end{figure*}


\begin{abstract}
Learning robot manipulation policies from human video demonstrations
constitutes a promising avenue for scalable robot learning. However,
comparing different human-to-robot (H2R) transfer methods 
remains challenging,
as existing approaches are evaluated under different settings, including differing
task suites, scene layouts, object instances, and amounts of robot
supervision. To address this challenge, we present \textbf{H2RBench}, a Real2Sim benchmark for evaluating H2R transfer methods.
\textbf{H2RBench} provides a standardized protocol built on
real human video demonstrations and simulated robot demonstrations, and
includes four manipulation tasks spanning diverse interaction requirements.
We evaluate multiple representative H2R transfer methods, each adopting a different strategy for bridging the embodiment gap.
Using H2RBench, we systematically characterize
how each method scales with the amount of human demonstrations,
revealing that methods differ substantially in their ability to
leverage additional human data. We further show that simulation performance is broadly predictive of
real-world robot performance, with an overall Pearson correlation of $r=0.89$, Spearman correlation of $\rho=0.85$ and and Mean Maximum Rank Violation (MMRV) of $0.06$ across method--task configurations. These results establish H2RBench as a practical and scalable benchmark for comparative H2R evaluation prior to real-world deployment. Videos and code are available on our
\href{https://davidz2851.github.io/h2rbench.github.io/}{project website}.

\end{abstract}



\section{Introduction}

Learning robot manipulation policies at scale remains fundamentally
bottlenecked by the cost of collecting robot demonstrations.
Large-scale robot data collection is constrained by hardware cost,
teleoperation throughput, and the difficulty of unifying data across
diverse robot platforms, control interfaces, and action spaces~\cite{o2024open,khazatsky2024droid}.
Compared with teleoperated robot demonstrations, human video
demonstrations offer a promising alternative~\cite{kareer2025emergence, nair2022r3m, bahl2023affordances},
as they can be collected with off-the-shelf cameras, without robot
hardware, and at the natural speed of human manipulation, making them
substantially faster and cheaper to scale per task~\cite{kareer2025egomimic, xiong2022robotube, liu2025egozerorobotlearningsmart}.
This opportunity has motivated a growing body of work on task-specific
human-to-robot (H2R) transfer~\cite{xu2023xskill, wang2023mimicplay,
ren2025motion, haldar2025point, lepert2025phantom,
collins2025amplify, krishna2026ghost}, which leverages third-person
human demonstration videos together with limited robot supervision to
train downstream robot manipulation policies.
As H2R methods proliferate, a fundamental question remains
underexplored: \emph{how should we systematically evaluate and compare
them?}

Answering this question is non-trivial for several reasons.
First, existing H2R methods are evaluated on heterogeneous task sets
with varying object types, scene configurations, and task difficulties,
making cross-method comparison unreliable~\cite{wang2023mimicplay, haldar2025point,lepert2025phantom}. Second, these methods differ not only in their technical approach, but
also in the amount of robot demonstration data they consume, conflating
methodological differences with data efficiency differences.
Third, real-robot evaluation is expensive, time-consuming, and difficult to reproduce at scale~\cite{khazatsky2024droid}. 

While simulation offers a scalable alternative, building a meaningful H2R benchmark in simulation is itself challenging, particularly on the human side of the interaction. Most existing H2R transfer methods are designed for human demonstrations captured in the real world: their hand and arm detection front-ends, such as MANO-based hand mesh estimators, are themselves trained on real human imagery~\cite{pavlakos2024reconstructing}. 
However, reproducing human interaction in simulation is non-trivial, as simulated hands differ from real hands both visually and physically. Existing approaches introduce additional artifacts: replaying recorded hand motion requires hand-model fitting and actuation, leading to reconstruction and contact errors~\cite{xu2025dexcanvas}, while direct teleoperation introduces retargeting artifacts~\cite{qin2022one}. As a result, evaluation on simulated hands may not faithfully reflect real-world H2R performance.
Moreover, the value of a simulation benchmark hinges on whether
relative policy performance in simulation is predictive of real-world
performance. Prior work has established this property for robot-only
policies---SIMPLER~\cite{li2024evaluating},
PolaRiS~\cite{jain2025polaris}, and RobotArena~$\infty$~\cite{jangir2025robotarena}
show that simulation provides a reliable signal of relative
real-world policy performance for policies trained on robot
demonstrations---but whether the same holds for H2R methods remains
unverified.

We address these gaps with \textbf{H2RBench}, a real-to-sim benchmark for human-to-robot transfer.
We
directly use real-world human videos as input to all evaluated
methods, mirroring how H2R methods are deployed in practice~\cite{wang2023mimicplay, lepert2025phantom, haldar2025point,
ren2025motion, collins2025amplify}. Each task environment is reconstructed from real-world objects and scenes via a real-to-sim pipeline~\cite{jain2025polaris}. We vary
the amount of human demonstrations to characterize how each method scales
with human data. Finally, we validate the benchmark by comparing relative policy
performance in simulation and on the real robot.

Our evaluation yields several key findings.
First, for current human-to-robot methods, human demonstrations most consistently improve high-level
interaction behaviors, while precise
late-stage control in contact-rich manipulation remains significantly
more challenging.
Second, H2R methods differ substantially in how effectively they
leverage additional human demonstrations: some scale consistently
with increasing human data, while others exhibit only limited gains.
Third, we show that simulation performance is broadly predictive of
relative real-world policy performance for H2R methods, supporting
the use of H2RBench as a practical proxy for real-world evaluation.

In summary, we make the following contributions:
\begin{itemize}
    \item A real-to-sim benchmark for H2R transfer consisting of manipulation
    tasks reconstructed from real-world scenes, spanning diverse precision
    requirements, contact dynamics, and task horizons, with task-specific
    human and robot demonstrations collected for each task.

    \item Empirical insights into how H2R methods scale with increasing amounts of human demonstration data, and a taxonomy of failure modes that informs future H2R algorithm design.

    \item An empirical Sim-vs.-Real validation showing that simulation
    performance is broadly predictive of relative real-world policy
    performance for H2R methods.
\end{itemize}


\section{Related Work}

\textbf{Robot Manipulation Benchmarks and Sim-vs.-Real Evaluation.}
A large family of manipulation benchmarks evaluates policies trained
from robot demonstrations, ranging from broad task suites~\cite{yu2020meta, james2020rlbench, gu2023maniskill2,
tao2024maniskill3, heo2025furniturebench} to
imitation-learning-focused protocols~\cite{mandlekar2021matters, liu2023libero}.
More recent work studies whether simulation can serve as a reliable proxy for evaluating real-world robot policies.
SIMPLER~\cite{li2024evaluating} introduced this paradigm by constructing
simulated counterparts of real robot setups and showing strong
correlation between simulation and real-world policy performance.
PolaRiS~\cite{jain2025polaris} further scales this idea through
high-fidelity neural reconstruction of real-world scenes, while
RobotArena~$\infty$~\cite{jangir2025robotarena} studies scalable
automatic evaluation in generated real-to-sim environments.
However, these benchmarks focus exclusively on robot-only policies and
do not evaluate methods that use human videos as a primary source of
supervision.

Among prior benchmarks, RoboTube~\cite{xiong2022robotube} is the
most closely related to H2RBench, as it collects human
videos together with simulated environments for robot learning from
human demonstrations. However, RoboTube primarily focuses on
reward-learning approaches that leverage human videos within online reinforcement learning settings; in contrast, we focus on 
H2R transfer methods trained with limited robot
data and more human data. Consequently, RoboTube does not standardize
robot-demonstration budgets or explicitly vary the amount of human
supervision across methods, making it difficult to systematically
measure how effectively different H2R transfer methods leverage
additional human demonstrations.
Further, H2RBench studies
whether simulation performance reliably predicts real-world performance for
H2R methods, providing the first systematic Sim-vs.-Real validation
in the H2R setting. A broader comparison between H2RBench and prior real-to-sim manipulation benchmarks is summarized in Table~\ref{tab:benchmark-comparison}.

\begin{table}[h]
\centering
\caption{Comparison of real-to-sim robot manipulation benchmarks across key benchmark properties for H2R evaluation. \textbf{H2R}: human-to-robot transfer evaluation; \textbf{Quant. Sim-vs.-Real}: quantitative sim-vs.-real performance validation; \textbf{Multi-stage SR}: stage-wise success metrics.
}
\label{tab:benchmark-comparison}

\resizebox{\columnwidth}{!}{%
\begin{tabular}{lccccc}
\toprule
\textbf{Benchmark}
& \textbf{Primary Focus}
& \textbf{H2R}
& \textbf{Quant. Sim-vs.-Real}
& \textbf{Multi-stage SR} \\
\midrule

SIMPLER~\cite{li2024evaluating}
& Sim-real correlation for pretrained policies
& \xmark  & \cmark & \xmark \\

PolaRiS~\cite{jain2025polaris}
& Neural real-to-sim reconstruction
& \xmark  & \cmark & \cmark \\

RobotArena~$\infty$~\cite{jangir2025robotarena}
& Scalable evaluation of generalist policies
& \xmark & \cmark & \cmark \\

RoboTube~\cite{xiong2022robotube}
& Human-video reward learning
& \cmark  & \xmark & \xmark \\

\midrule

\textbf{H2RBench (ours)}
& Controlled H2R evaluation with human data scaling
& \cmark  & \cmark & \cmark \\

\bottomrule
\end{tabular}%
}
\end{table}

\textbf{Human-to-Robot Policy Transfer.}
Learning robot manipulation policies from human video demonstrations
has emerged as a promising paradigm for robot learning. However,
transferring behavior from human videos to robot policies remains
challenging due to the embodiment gap between humans and robots.
Prior work has explored using human videos for robot learning through
visual representation pretraining~\cite{nair2022r3m,
radosavovic2023real}, reward learning~\cite{sermanet2018time,
zakka2022xirl}, and extracting affordances or interaction
priors~\cite{bahl2023affordances,mendonca2023structured}. More recent H2R methods directly address the embodiment gap between human demonstrations and robot execution through different modeling choices and training strategies~\cite{lepert2025phantom,haldar2025point,collins2025amplify,krishna2026ghost,wang2023mimicplay}.
While these methods demonstrate promising results, they are evaluated
on different tasks, with different amounts of robot supervision, making direct comparison
difficult. H2RBench provides a unified evaluation framework for fair
and reproducible comparison across H2R transfer methods under
standardized data and evaluation settings.


\section{H2RBench}

\subsection{Overview}

\begin{figure*}[t]
    \centering
    \includegraphics[width=\textwidth]{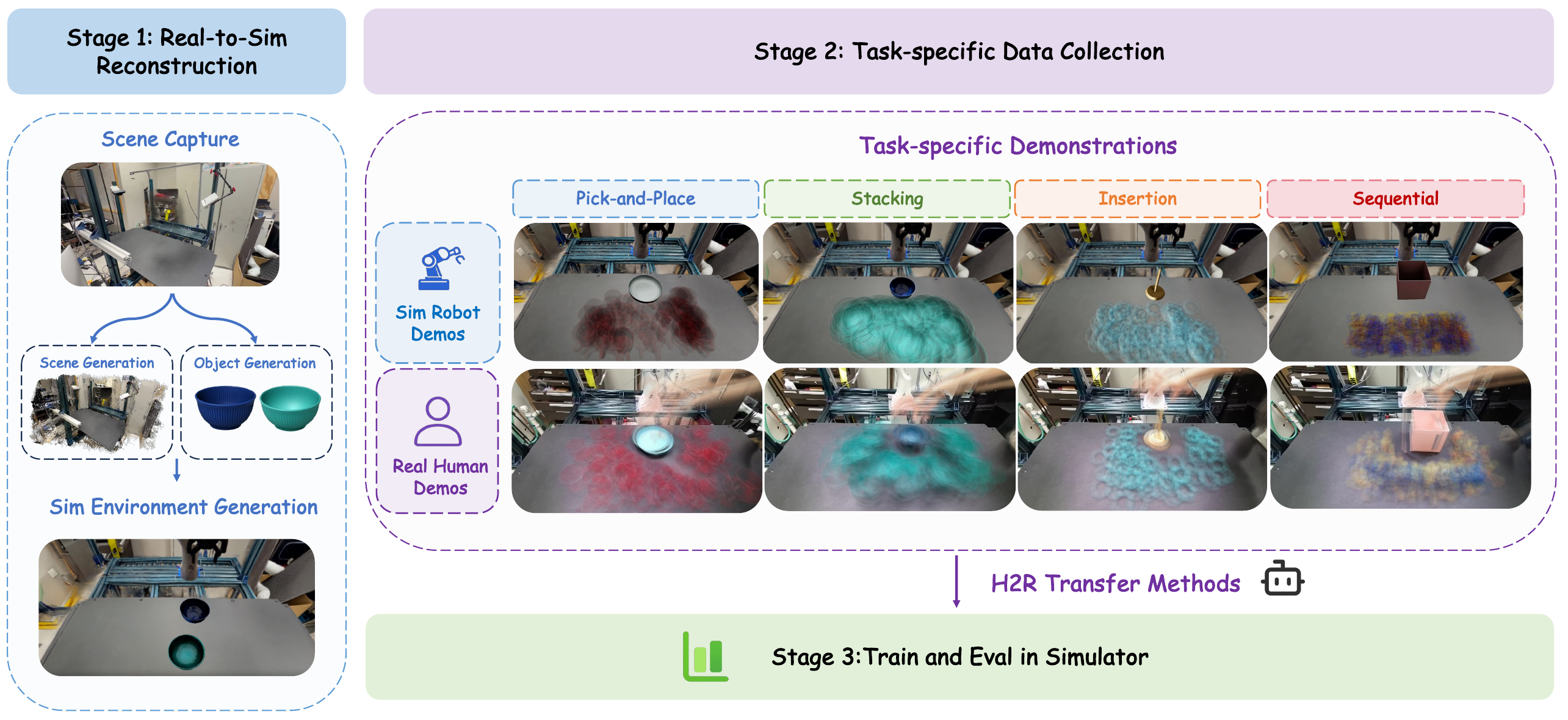}
    \caption{
    \textbf{Overview of the H2RBench evaluation pipeline.}
We reconstruct real-world workspaces into Isaac Lab simulation environments
and collect task-specific human and robot demonstrations across four
manipulation tasks with diverse object placement distributions.
Methods are then trained and evaluated in simulation under a shared
benchmark protocol for controlled H2R comparison.}
    \label{fig:h2rbench_pipeline}
\end{figure*}

\textbf{H2RBench}
provides a standardized benchmark for evaluating task-specific
H2R transfer methods that use human video
demonstrations together with limited robot demonstrations to learn
robot manipulation policies.
The benchmark consists of four manipulation tasks, each implemented
as a simulation environment reconstructed from a real-world scene
via a real-to-sim pipeline.
All baselines are trained on the same dataset of real human video
demonstrations together with simulated robot demonstrations under a shared training and evaluation protocol.
Two design choices distinguish H2RBench from existing
manipulation benchmarks. First, it targets methods that learn from
human videos rather than robot demonstrations alone. Second, we hold
the robot demonstration budget fixed across methods while varying the
number of human demonstrations, enabling controlled analysis of how
different methods scale with human supervision—a core property of H2R
transfer that prior benchmarks do not explicitly measure.

\textbf{H2RBench} provides a unified observation-action interface while supporting diverse H2R policy representations. Observations include synchronized multi-view RGB-D images, robot proprioception, and language; human demonstrations use front and side views, while robot demonstrations additionally include wrist-mounted observations. Actions are represented in the robot base frame as 3D translation, 6D rotation~\cite{zhou2019continuity}, and a binary gripper command, following prior H2R methods~\cite{lepert2025phantom, haldar2025point}.

\subsection{Task Suite}

We design four tasks that cover a broad range of manipulation tasks requiring varying levels of required precision and task horizon.
Table~\ref{tab:tasks} summarizes the key properties of each task.

\begin{table}[h]
\centering
\caption{Overview of the four manipulation tasks in H2RBench,
including task horizon, precision requirement, the fixed number
of robot demonstrations, and the maximum number of human
demonstrations used in scaling experiments.}
\label{tab:tasks}
\resizebox{\columnwidth}{!}{%
\begin{tabular}{lcccc}
\toprule
\textbf{Task} & \textbf{Horizon} & \textbf{Precision Requirement} & \textbf{Robot Demos} & \textbf{Max Human Demos} \\
\midrule

Pick-and-Place (Mug $\rightarrow$ Plate)
& Short
& Coarse
& 40
& 100 \\

Stacking (Bowl $\rightarrow$ Bowl)
& Short
& \cellcolor{skyblue1} Moderate
& 100
& 300 \\

Insertion (Donut $\rightarrow$ Peg)
& Short
& \cellcolor{skyblue2} Precise
& 100
& 300 \\

Sequential Manipulation (Blocks $\rightarrow$ Box)
& \cellcolor{skyblue2} Long
& \cellcolor{skyblue1} Moderate
& 100
& 300 \\

\bottomrule
\end{tabular}%
}
\end{table}

\subsection{Real-to-Sim Reconstruction and Data Collection}

\textbf{Scene Reconstruction.} As illustrated in Fig.~\ref{fig:h2rbench_pipeline}, H2RBench consists of three stages: real-to-sim reconstruction, task-specific data collection, and simulation-based training and evaluation. All task environments are built with PolaRiS~\cite{jain2025polaris} on top of NVIDIA Isaac Lab~\cite{mittal2025isaac}. For scene reconstruction, we replace the original PolaRiS pipeline with Marble, which produces higher-quality 3D meshes and Gaussian Splatting reconstructions, resulting in more visually consistent and geometrically faithful simulation assets. Interactive objects are reconstructed from multiview reference images using TRELLIS~\cite{xiang2025structured}, then imported into the simulator and scaled to match their real-world dimensions.

\textbf{Data Collection.}
For each task, we collect human and robot demonstrations.
Human demonstrations are recorded as third-person RGB videos from
multiple operators performing the task in the real-world workspace
using fixed camera viewpoints consistent with the reconstructed
simulation scene. Robot demonstrations are collected in both simulation
and the real world with an extra wrist camera. In simulation, trajectories are generated using the
cuRobo motion planner~\cite{sundaralingam2023curobo}. In the real
world, demonstrations are collected using a Franka Emika Panda arm via
the GELLO teleoperation interface~\cite{wu2024gello}. Additional
details on the data collection setup are provided in
Sec.~\ref{sec:supp_data_collection}.

\subsection{Evaluation Protocol}
\label{sec:eval-protocol}

\textbf{Evaluation Metric.}
We evaluate each policy using a multi-stage success rate, where each
task is decomposed into cumulative stages reflecting task progress.
We report both final-stage success rate and the average success rate
across all stages as an overall measure of task completion.

\textbf{Sim-vs.-Real Validation.}
For each task and each method, we additionally train a separate policy
using a real-world training pipeline consisting of real human
demonstrations together with real robot demonstrations, and evaluate
the resulting policy on the physical robot. We then compare the
rankings of methods obtained under the simulation pipeline and the
real-world pipeline using rank-correlation metrics (Pearson $r$,
Spearman $\rho$, Mean Maximum Rank Violation (MMRV)). This evaluation
quantifies whether H2RBench provides a reliable proxy for real-world
H2R method comparison.



\section{Methods Evaluated}

\textbf{Phantom (Visual Representation Alignment).}
Phantom~\cite{lepert2025phantom} edits human demonstrations to
visually align them with robot observations by inpainting the human
arm and overlaying a rendered robot arm before imitation learning.

\textbf{Point Policy (Keypoint-Based Representation).}
Point Policy~\cite{haldar2025point} represents human and robot
behavior using shared 3D keypoints extracted from multi-view
observations and predicts future robot keypoints directly in this
representation space.

\textbf{AMPLIFY (Latent Dynamics Representation).}
AMPLIFY~\cite{collins2025amplify} decouples interaction dynamics
modeling from action inference by first learning a latent dynamics
representation from videos and then predicting robot actions from
that representation using robot demonstrations.

\textbf{GHOST (Hierarchical Subgoal Representation).}
GHOST~\cite{krishna2026ghost} decomposes H2R transfer into a
high-level subgoal planner learned from human and robot
demonstrations and a low-level robot controller trained to execute
those subgoals from robot demonstrations.


\section{Experiments}

We use H2RBench to systematically evaluate human-to-robot (H2R)
transfer methods.
Beyond comparing final task performance, our goal is to characterize
how human supervision interacts with different embodiment-bridging
strategies across manipulation regimes with varying precision
requirements and horizon length. Our experiments are organized around three questions:

\textbf{Q1: Capability and human transfer effectiveness.}
    How do H2R methods compare in overall task performance, and to
    what extent do human demonstrations improve or degrade execution
    relative to robot-only training?

\textbf{Q2: Scaling with human supervision.}
    How does policy performance evolve as the amount of human
    supervision increases?

\textbf{Q3: Sim-vs.-Real fidelity.}
    Does evaluation in H2RBench reliably predict real-world policy performance?


\subsection{Capability and Human Transfer Effectiveness (Q1)}

\textbf{Setup.}
Unless otherwise specified, all methods use the recommended
hyperparameters from their original publications and are evaluated
under the H2RBench protocol in
Section~\ref{sec:eval-protocol} over 90 randomized rollouts across 3 seeds. To quantify the
effect of human supervision, we compare robot-only and robot+human
training under identical architectures and hyperparameters. We report both average task progression (\textbf{Prog.}) and
final-stage success rate (\textbf{SR}). Since some methods already
perform strongly under robot-only training, absolute success-rate
improvements may not fully reflect the benefit of human
demonstrations. We therefore additionally report the log-odds
transfer gain:

\[
\Delta_{\text{logit}}
=
\log\frac{p_{R+H}}{1-p_{R+H}}
-
\log\frac{p_R}{1-p_R},
\]

where $p_R$ and $p_{R+H}$ denote the success rates under robot-only
and robot+human training, respectively. Compared with raw
success-rate differences, this metric better accounts for gains near
performance saturation and provides a more balanced measure of human
transfer effectiveness.

\textbf{Results.}
Table~\ref{tab:multiseed_main_1} compares overall task performance across methods
under full human supervision.
Overall, GHOST~\cite{krishna2026ghost} achieves the strongest and
most consistent performance across tasks, while AMPLIFY~\cite{collins2025amplify} performs particularly
well on stacking and Point Policy~\cite{haldar2025point} remains competitive on
pick-and-place and stacking.
Across methods, approaches leveraging hierarchical decomposition or
explicit subgoal representations tend to achieve stronger and more
robust performance, suggesting that structured intermediate
abstractions are effective for bridging embodiment differences in
manipulation. More detailed qualitative analysis is provided in Appendix~\ref{sec:failure_case_analysis}.

Fig.~\ref{fig:human_gain} further shows that the benefit of human
supervision varies substantially across both tasks and methods.
Human demonstrations consistently improve performance on pick-and-place
and remain beneficial for stacking, indicating that current H2R
methods effectively transfer high-level spatial intent.
In contrast, performance often degrades for insertion, while improvements are more limited for sequential manipulation, where fine-grained contact dynamics and error accumulation across repeated interaction steps remain challenging.  Overall, while human demonstrations generally improve task
performance, H2R methods differ substantially in how effectively they
leverage additional human data, highlighting the importance of
evaluating both overall performance and transfer effectiveness in
H2R benchmarks.

\begin{figure}[t]
    \centering

    \begin{subfigure}[t]{0.49\columnwidth}
        \centering
        \includegraphics[width=\linewidth]{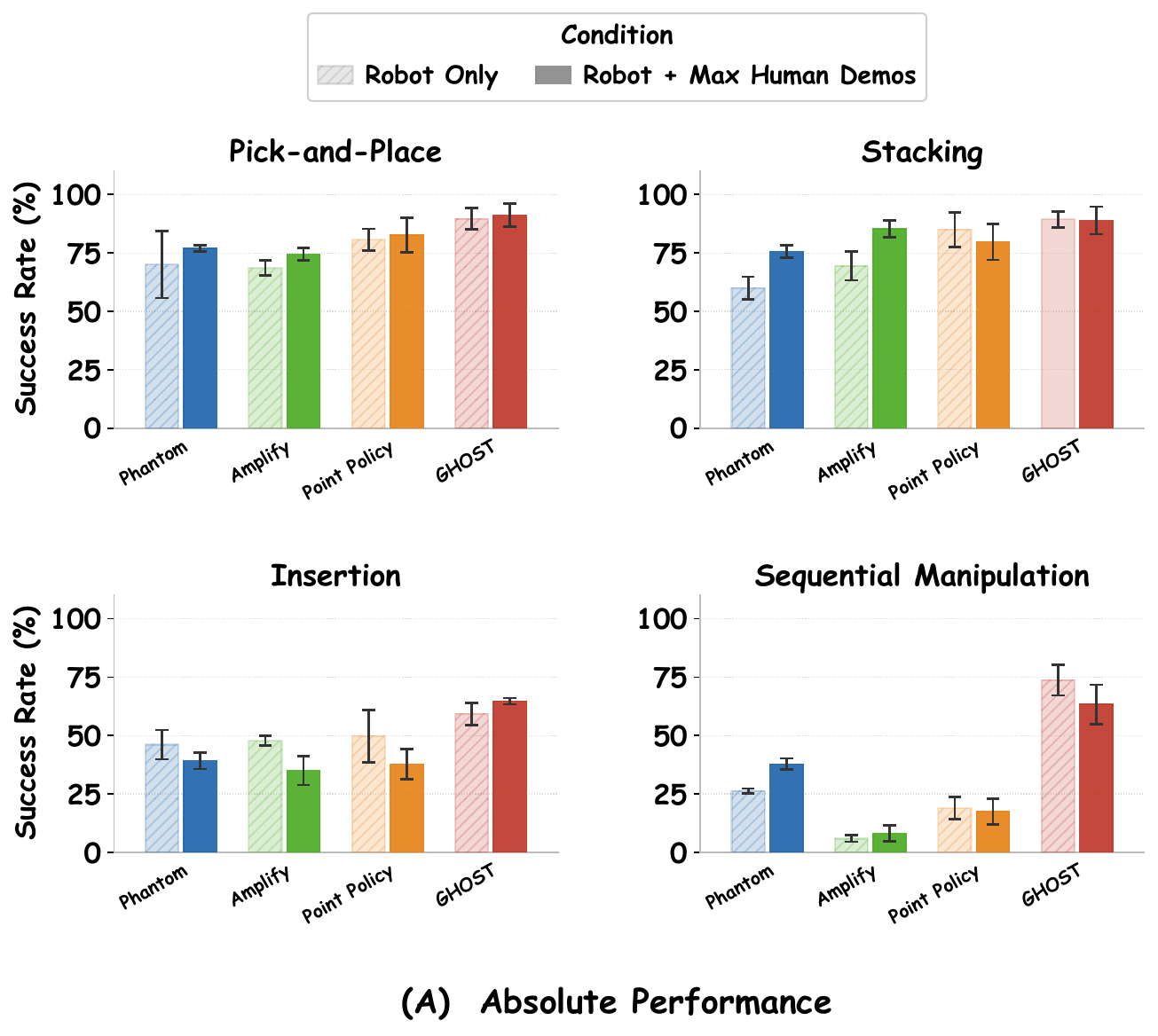}
        \label{fig:absolute_perf}
    \end{subfigure}
    \hfill
    \begin{subfigure}[t]{0.49\columnwidth}
        \centering
        \includegraphics[width=\linewidth]{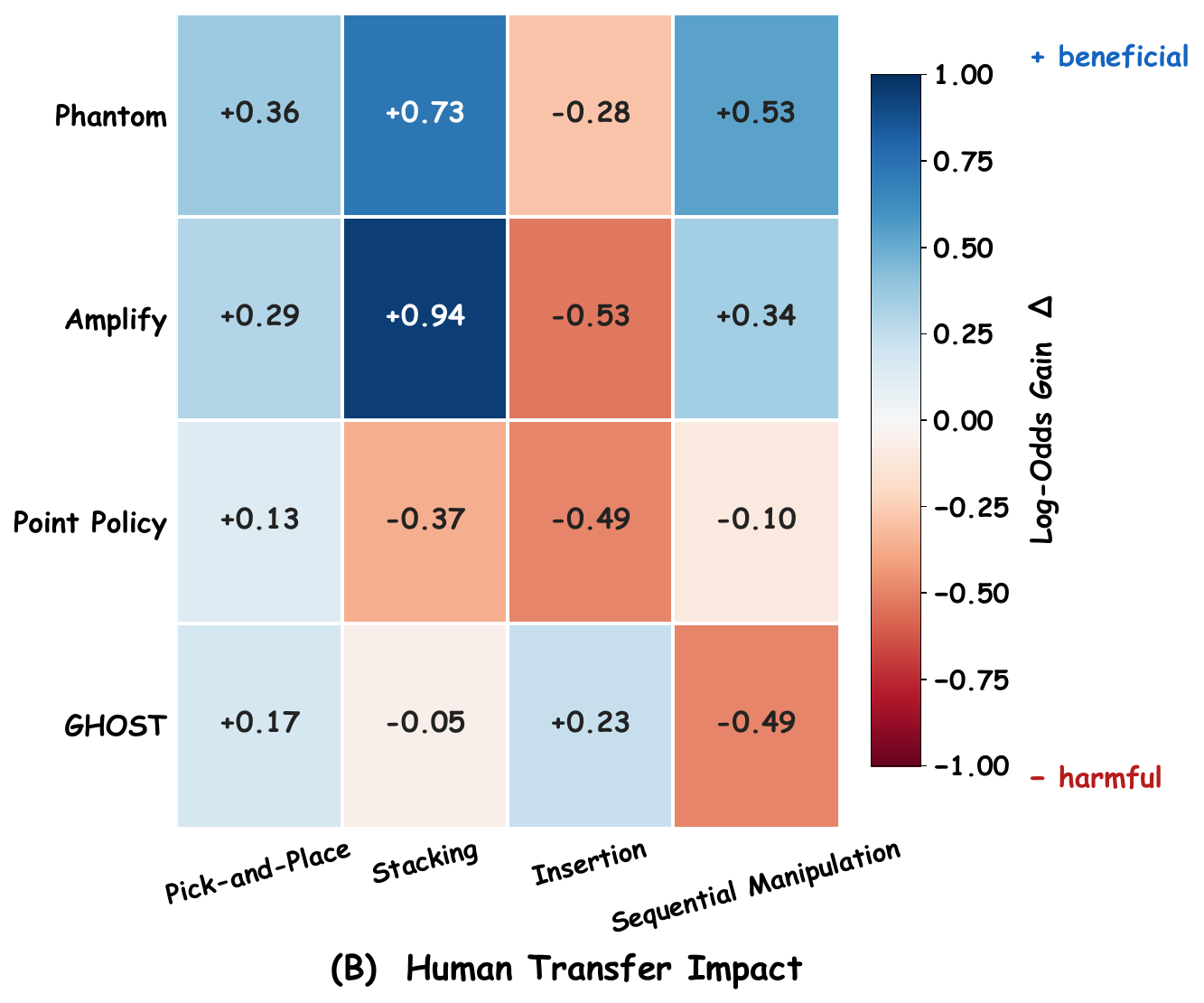}
        \label{fig:log_gain}
    \end{subfigure}

    \caption{
    Capability and human transfer impact across H2R methods and tasks.
    \textbf{Left}: absolute task performance.
    \textbf{Right}: human transfer impact (log-odds gain) 
    }
    \label{fig:human_gain}
\end{figure}

\begin{table}[h]
\centering
\caption{
Multi-seed benchmark results under robot + maximum human demonstration
training. We report mean and standard deviation across three random seeds
for task progression (\textbf{Prog.}) and final-stage success rate
(\textbf{SR}).
}
\label{tab:multiseed_main_1}

\scriptsize
\setlength{\tabcolsep}{3.5pt}
\renewcommand{\arraystretch}{0.95}

\resizebox{\columnwidth}{!}{%
\begin{tabular}{lcccccccc}
\toprule
\multirow{2}{*}{\textbf{Method}}
& \multicolumn{2}{c}{\textbf{Pick-and-Place}}
& \multicolumn{2}{c}{\textbf{Stacking}}
& \multicolumn{2}{c}{\textbf{Insertion}}
& \multicolumn{2}{c}{\textbf{Sequential Manipulation}} \\
\cmidrule(lr){2-3}
\cmidrule(lr){4-5}
\cmidrule(lr){6-7}
\cmidrule(lr){8-9}
& \textbf{Prog.} & \textbf{SR}
& \textbf{Prog.} & \textbf{SR}
& \textbf{Prog.} & \textbf{SR}
& \textbf{Prog.} & \textbf{SR} \\
\midrule

Phantom~\cite{lepert2025phantom}
& 77.0 $\pm$ 1.4 & 56.7 $\pm$ 2.7
& 75.6 $\pm$ 2.7 & 25.6 $\pm$ 6.8
& \underline{39.2} $\pm$ 3.6 & \underline{7.8} $\pm$ 4.2
& \underline{37.8} $\pm$ 2.4 & \underline{13.3} $\pm$ 4.7 \\

AMPLIFY~\cite{collins2025amplify}
& 74.4 $\pm$ 2.7 & 58.9 $\pm$ 4.2
& \underline{85.3} $\pm$ 3.7 & \underline{61.1} $\pm$ 9.6
& 35.0 $\pm$ 6.2 & 2.2 $\pm$ 1.6
& 8.1 $\pm$ 3.4 & 0.0 $\pm$ 0.0 \\

Point Policy~\cite{haldar2025point}
& \underline{82.6} $\pm$ 7.4 & \underline{62.2} $\pm$ 11.7
& 79.7 $\pm$ 7.7 & 61.1 $\pm$ 17.1
& 37.8 $\pm$ 6.5 & 6.7 $\pm$ 11.5
& 17.4 $\pm$ 5.6 & 0.0 $\pm$ 0.0 \\

GHOST~\cite{krishna2026ghost}
& \textbf{91.1} $\pm$ 4.9 & \textbf{82.2} $\pm$ 6.9
& \textbf{88.9} $\pm$ 5.9 & \textbf{76.7} $\pm$ 8.8
& \textbf{64.7} $\pm$ 1.3 & \textbf{34.5} $\pm$ 6.9
& \textbf{63.3} $\pm$ 8.4 & \textbf{38.9} $\pm$ 15.0 \\

\bottomrule
\end{tabular}
}
\end{table}














\subsection{Scaling with Human Supervision (Q2)}

\textbf{Setup.}
We
vary the number of human demonstrations while keeping the robot
demonstration budget fixed. We evaluate scaling trends using average task progression
(\textbf{Prog.}) success rate.

\begin{figure}[t]
    \centering
    \includegraphics[width=\columnwidth]{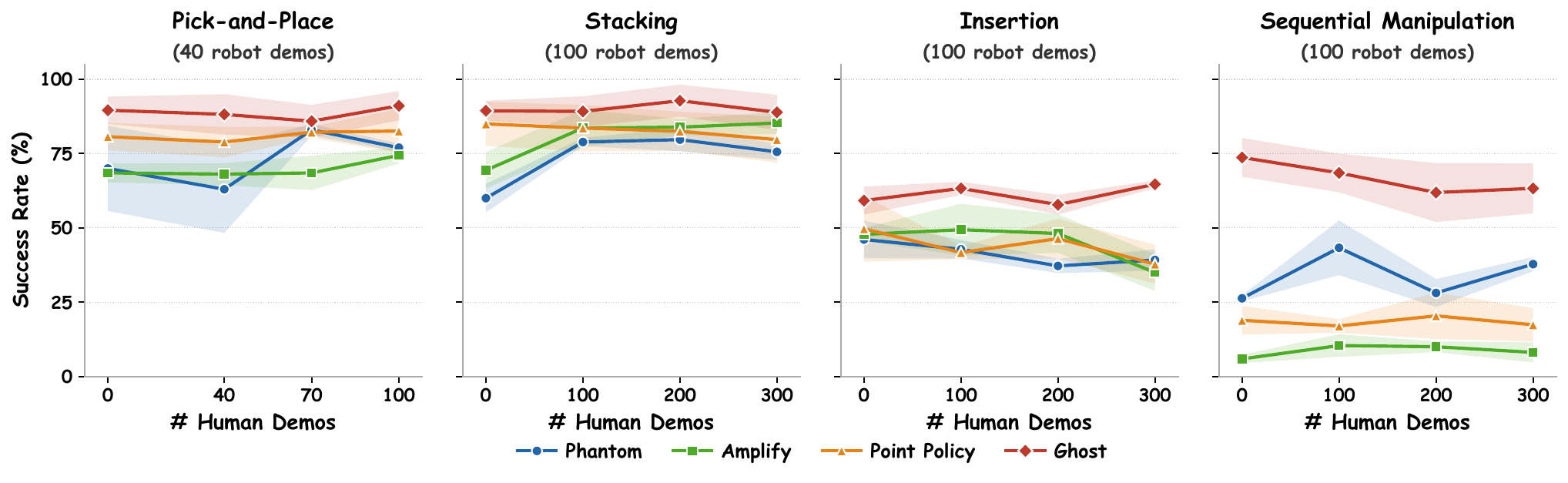}
    \caption{
    Scaling performance with respect to the number of human demonstrations.
    }
    \label{fig:scaling}
\end{figure}

\textbf{Results.}
Figure~\ref{fig:scaling} shows scaling performance as a function of
the number of human demonstrations for each task.
For simpler manipulation tasks, most methods
improve steadily as additional human demonstrations are added,
suggesting that current H2R approaches can effectively leverage human
video to improve coarse transport and placement behavior.
In contrast, scaling trends become less consistent for more
precision-sensitive tasks.
Stacking exhibits strong method-dependent behavior, with some methods
continuing to improve while others plateau.
Insertion further amplifies this effect, where increasing human
demonstrations often provides limited or unstable gains.
Sequential manipulation reveals diminishing returns under long-horizon
execution.
While some methods scale well with more human data,
others saturate or fluctuate with increased supervision,
suggesting that compounding execution errors remain difficult to
correct through additional human demonstrations alone. Overall, these results suggest that the value of scaling human
demonstrations depends strongly on both task structure and the
embodiment-bridging strategy used by the H2R method.

\subsection{Sim-vs.-Real Fidelity (Q3)}
\label{sec:exp-q3}

\textbf{Setup.} For each method-task pair, we train:
(i) a \emph{simulation-pipeline} ($\pi_\text{sim}$) policy using real human
demonstrations and simulated robot demonstrations, and
(ii) a \emph{real-world-pipeline} ($\pi_\text{real}$) policy using real human
demonstrations and real robot demonstrations. $\pi_\text{sim}$ is evaluated in the reconstructed
environment, while $\pi_\text{real}$ is
evaluated on the physical robot.
We then measure Sim-vs.-Real correspondence between simulated and real-world success rates.

\begin{wrapfigure}[22]{r}{0.48\columnwidth}
    \vspace{-10pt}
    \centering

    \footnotesize
\setlength{\tabcolsep}{3pt}
\renewcommand{\arraystretch}{0.95}

\captionof{table}{
Sim-vs.-Real fidelity metrics under \textbf{Robot + Max Human Demos}.
}
\label{tab:sim2real}

\resizebox{0.42\columnwidth}{!}{%
\begin{tabular}{ccc}
\toprule
\textbf{Pearson $r$ $\uparrow$}
& \textbf{Spearman $\rho$ $\uparrow$}
& \textbf{MMRV $\downarrow$} \\
\midrule
0.894 & 0.851 & 0.06 \\
\bottomrule
\end{tabular}
}

    \vspace{8pt}

    \includegraphics[
        width=0.46\columnwidth,
        keepaspectratio
    ]{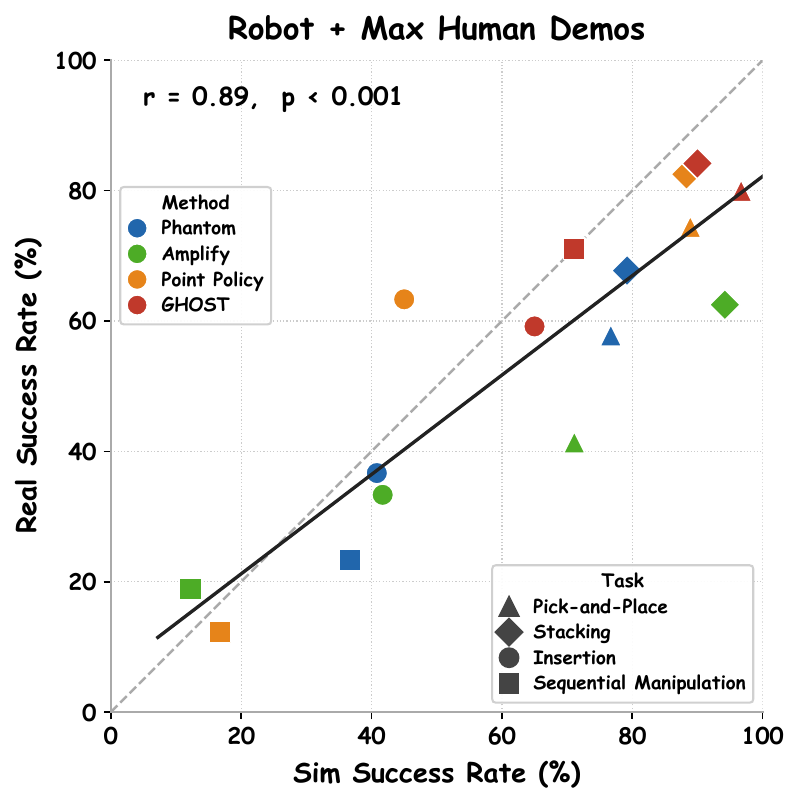}

    \captionof{figure}{
    Sim-vs.-Real correlation between simulated and real-world policy performance.
    }
    \label{fig:sim-real}

    \vspace{-8pt}
\end{wrapfigure}

\textbf{Results.} Fig.~\ref{fig:sim-real} shows a strong correlation between simulated and real-world policy performance across method–task pairs. Table~\ref{tab:sim2real} further shows high Pearson $r$ and Spearman $\rho$, together with a low Mean Maximum Rank Violation (MMRV) of 0.06, indicating that simulation reliably preserves both performance trends and method rankings. Overall, these results support \textbf{H2RBench} as a practical proxy for H2R evaluation, enabling efficient benchmarking in simulation while reducing the cost of real-world experimentation.

\section{Limitations}

H2RBench has several limitations. First, it currently focuses on
tabletop manipulation with primarily rigid objects and does not yet
cover deformable manipulation, articulated objects, or bimanual
coordination. Second, H2RBench evaluates H2R methods using task-specific human
demonstrations collected under controlled third-person camera setups;
extending benchmark evaluation to egocentric video or human
demonstrations collected ``in-the-wild'' remains an
important direction for future work.
Finally, H2RBench inherits
limitations of current real-to-sim reconstruction pipelines,
including imperfect physical fidelity and limited support for complex
dynamics.

\section{Conclusion}

We introduced \textbf{H2RBench}, a real-to-sim benchmark for
evaluating human-to-robot transfer under a unified evaluation
framework. Across four manipulation tasks and representative H2R methods, our
experiments show that the benefit of human demonstrations is highly
task- and method-dependent: current H2R methods transfer high-level
interaction behavior more reliably than precise control.
We further show that \textbf{H2RBench} is broadly
predictive of relative real-world policy performance, supporting the
use of real-to-sim environments as a practical proxy for comparative
H2R evaluation. 

\clearpage
\acknowledgments{
This material is based upon work supported by the Toyota Research Institute. David Held holds concurrent appointments as a Professor at CMU and as an Amazon Scholar. This paper describes work performed at CMU and is not associated with Amazon. 
}


\bibliography{reference}


\clearpage
\appendix

\section*{Appendix}
\addcontentsline{toc}{section}{Appendix}

\section{Implementation Details}

\subsection{Detailed data collection pipeline}
\label{sec:supp_data_collection}
\textbf{Human Demonstrations.}
For each task, we collect real-world human demonstrations from
multiple operators performing the task in the reconstructed workspace.
H2RBench does not assume paired human-to-robot trajectories. Instead,
we evaluate whether H2R methods can transfer useful behavior priors
from human demonstrations under a shared observation-action interface.

\textbf{Robot Demonstrations.}
Robot demonstrations are collected in both simulation and the real
world. In simulation, we generate robot trajectories using the cuRobo
motion planner~\cite{sundaralingam2023curobo}. We use 40 demonstrations for
pick-and-place and 100 for the remaining tasks, and keep this robot
demonstration budget fixed across methods for fair comparison. In the
real world, demonstrations are collected using a Franka Emika Panda
arm through the GELLO teleoperation interface~\cite{wu2024gello}.

\subsection{Detailed Simulator Parameters}

Unless otherwise specified, all environments use the default physics
configuration from Isaac Lab / Isaac Sim. We make only minimal
task-specific adjustments to improve sim-to-real consistency.

In particular, we tune contact-related parameters such as object
friction and gripper fingertip friction for selected objects when
necessary. These adjustments primarily improve grasp stability and
contact behavior in tasks involving stacking or insertion, where small
physical mismatches can significantly affect success.

We avoid extensive per-task tuning and keep the simulator settings as
close as possible to the default Isaac Sim configuration to maintain a
consistent evaluation protocol across tasks.

\subsection{Detailed H2R Methods Implementation}
Unless otherwise noted, all H2R baselines are implemented following
the official codebases and default hyperparameter settings reported in
their original papers. To ensure fair comparison, we preserve each
method’s original architecture and training recipe whenever possible,
and introduce only minimal modifications necessary to adapt the method
to the H2RBench task setup and observation modalities. For hand pose estimation model, we just use the same model specified in the original paper. Below, we highlight several method-specific implementation details, particularly camera-view usage and preprocessing choices.

\begin{table}[h]
\centering
\caption{Camera views used during training for each H2R method.}
\label{tab:view_usage}

\footnotesize
\setlength{\tabcolsep}{8pt}
\renewcommand{\arraystretch}{1.0}

\begin{tabular}{lccc}
\toprule
\textbf{Method}
& \textbf{Front}
& \textbf{Left}
& \textbf{Wrist} \\
\midrule
Phantom & \cmark & \xmark & \xmark \\
Point Policy & \cmark & \cmark & \xmark \\
AMPLIFY & \cmark & \cmark & \xmark \\
GHOST & \cmark & \cmark & \cmark \\
\bottomrule
\end{tabular}
\end{table}
\subsubsection{Camera View Usage}
\label{sec:implementation_details_h2r}

Although all methods are evaluated under the same task suite and data
collection protocol, the camera views used during training differ
across methods due to differences in how each method bridges the
embodiment gap between human and robot demonstrations.
Table~\ref{tab:view_usage} summarizes the camera views used by each
evaluated method. In general, whether a camera view can be used
depends on whether human demonstrations can be consistently aligned
with robot observations in that perspective.

Phantom~\cite{lepert2025phantom} is trained using only the
\textbf{front view}, since its robot-inpainted human videos are
generated through end-effector retargeting and inverse kinematics,
which do not guarantee temporally consistent rendering across multiple
viewpoints. AMPLIFY~\cite{collins2025amplify} and Point
Policy~\cite{haldar2025point} use both \textbf{front-view} and
\textbf{left-view} observations, as they learn cross-embodiment
representations directly from shared human and robot videos without
requiring explicit rendered alignment. Neither method uses the robot
wrist camera because human demonstrations do not provide a paired
egocentric view. For AMPLIFY specifically, although the inverse
dynamics model is trained using robot-only demonstrations, it depends
on motion tracks predicted by upstream modules trained only on the
shared front-view and left-view videos, preventing consistent use of
wrist-camera observations. GHOST~\cite{krishna2026ghost}
additionally uses the robot \textbf{wrist camera} during low-level
policy training, since this stage is trained on robot-only
demonstrations and does not require paired human observations.

Overall, as shown in Table~\ref{tab:view_usage}, these differences
reflect an inherent trade-off in H2R learning: methods that require
explicit human--robot alignment are restricted to shared external
viewpoints, while hierarchical methods with robot-only downstream
control can additionally leverage robot-specific egocentric
observations such as the wrist camera.

\subsubsection{Practical Annotation Cost Analysis}

Beyond task performance, H2R methods also differ in
the amount of manual supervision required during data preprocessing
and annotation. Since annotation overhead directly affects the
scalability and practicality of H2R pipelines, we summarize the
manual labeling effort required by each evaluated method.

Phantom~\cite{lepert2025phantom},
AMPLIFY~\cite{collins2025amplify}, and GHOST~\cite{krishna2026ghost} primarily rely on automatic data
processing pipelines and require minimal manual annotation beyond
dataset collection. Point Policy~\cite{haldar2025point} requires
manual keypoint initialization on the first frame of a single
demonstration sequence per task, resulting in annotation effort that scales
linearly with the number of tasks. 

GHOST~\cite{krishna2026ghost} uses gripper-open close to automatically get subgoal key frames; for human demos, we compute the subgoal keyframes when the distance between the thumb and index fingers is less than a threshold for a given number of frames. The threshold is selected manually on a per-task basis.

\subsection{Task Details}
H2RBench includes four manipulation tasks with varying interaction
complexity. For each task, we define a language instruction and a
step-wise evaluation rubric used to compute task progression and
final success.

\begin{figure}[H]
    \centering
    \includegraphics[width=\columnwidth]{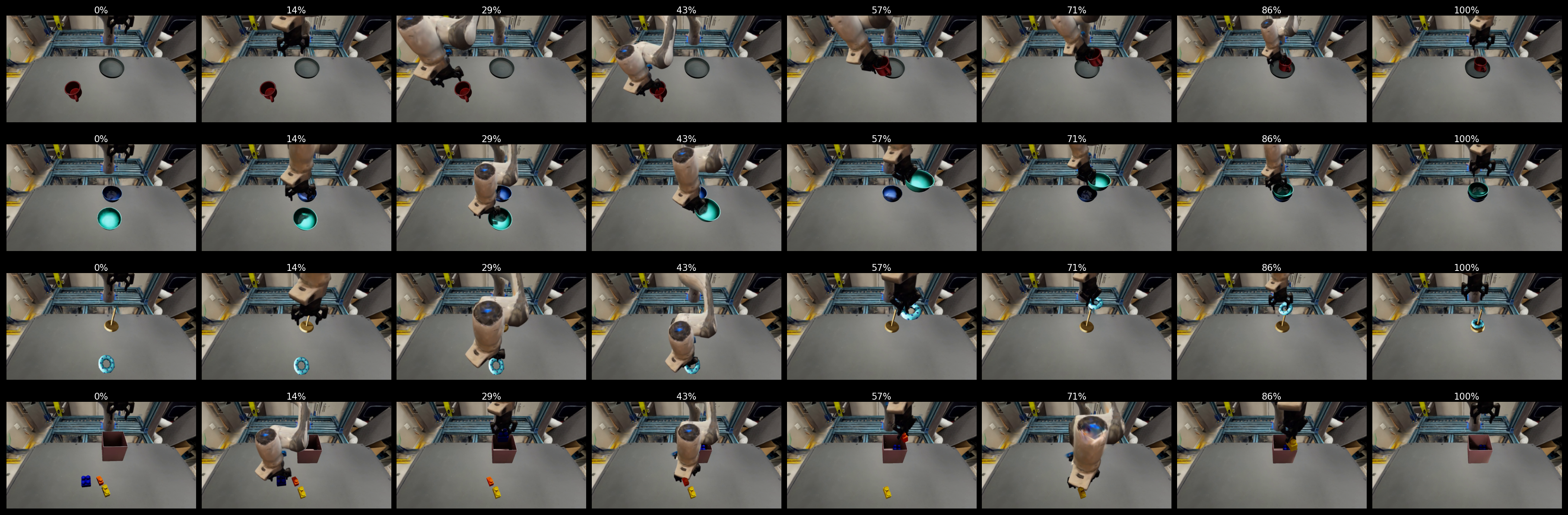}
    \caption{
    Visualization of the task progression for each task. We show representative intermediate
    states from the initial scene configuration to successful task
    completion. 
    }
    \label{fig:task_progression}
\end{figure}

Table~\ref{tab:tasks_details} summarizes the task instructions and evaluation
rubric for each task. Figure~\ref{fig:task_progression} shows
representative intermediate states during task execution from initial
configuration to successful completion. These visualizations are
illustrative and may contain finer-grained transitions than the
discrete stages used for evaluation.
\begin{table}[H]
\centering
\caption{
Task prompts and corresponding step-by-step evaluation rubrics for
the four H2RBench tasks. Scores are normalized between 0 and 1 for
comparability across environments.
}
\label{tab:tasks_details}

\footnotesize
\setlength{\tabcolsep}{6pt}
\renewcommand{\arraystretch}{1.0}

\begin{tabularx}{\columnwidth}{
p{2.8cm}
p{3.4cm}
X
}
\toprule
\textbf{Task} &
\textbf{Instruction} &
\textbf{Step-by-Step Rubric} \\
\midrule

Pick-and-Place
&
Put the mug onto the plate
&
\begin{enumerate}[nosep,leftmargin=1.2em]
\item Reach for the red cup
\item Lift the red cup
\item Place the red cup onto the blue plate
\end{enumerate}
\\
\midrule

Stacking
&
Stack the bowl onto the other bowl
&
\begin{enumerate}[nosep,leftmargin=1.2em]
\item Reach for the green bowl
\item Lift the green bowl
\item Reach for the blue bowl
\item Place the blue bowl on top of the green bowl
\end{enumerate}
\\
\midrule

Insertion
&
Insert the donut onto the peg
&
\begin{enumerate}[nosep,leftmargin=1.2em]
\item Reach for the blue donut
\item Lift the blue donut
\item Move the donut above the peg
\item Insert the donut onto the peg
\end{enumerate}
\\
\midrule

Sequential Manipulation
&
Pick up the toys and put them into the box
&
\begin{enumerate}[nosep,leftmargin=1.2em]
\item Place the blue toy into the pink box
\item Place the orange toy into the pink box
\item Place the yellow toy into the pink box
\end{enumerate}
\\

\bottomrule
\end{tabularx}
\end{table}

\section{Additional Results}
\subsection{Additional Stage-wise Transfer Analysis}

To localize where human demonstrations affect execution, we further
measure the stage-wise transfer impact:
\[
\Delta \mathrm{SR}_s
=
\mathrm{SR}^{R+H}_s
-
\mathrm{SR}^{R}_s,
\]
where $\mathrm{SR}^{R+H}_s$ and $\mathrm{SR}^{R}_s$ denote the
robot+human and robot-only success rates at stage $s$, respectively.

\begin{figure}[H]
    \centering
    \includegraphics[width=\columnwidth]{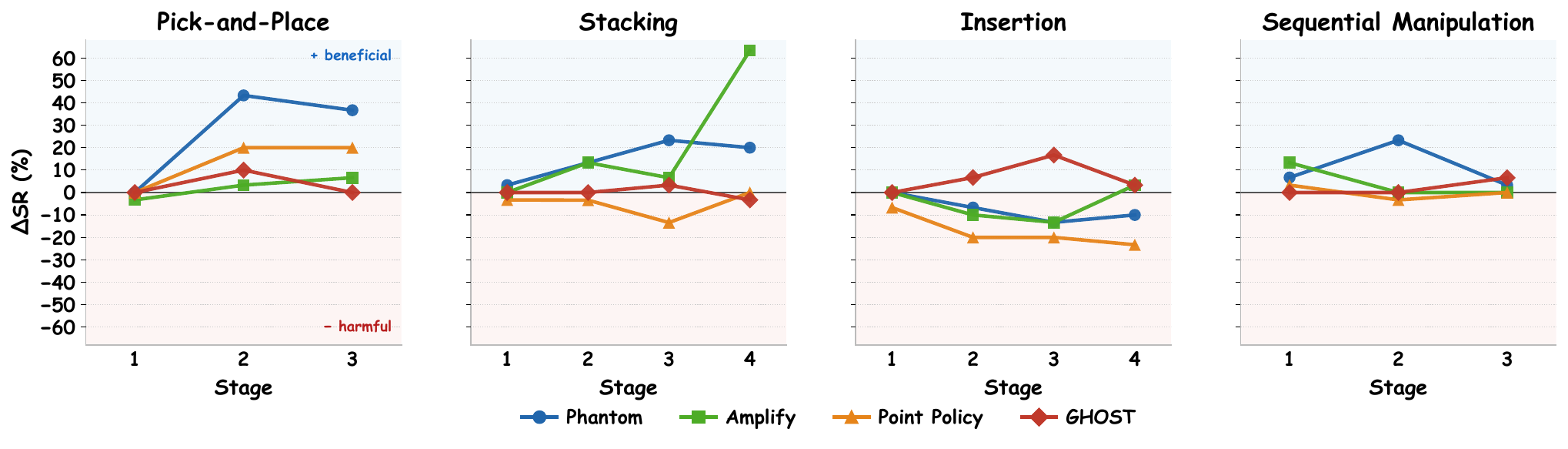}
    \caption{
    Stage-wise human transfer impact measured as
    $\Delta \mathrm{SR}_s =
    \mathrm{SR}^{R+H}_s - \mathrm{SR}^{R}_s$.
    Positive values indicate that adding human demonstrations improves
    stage completion, while negative values indicate degraded performance.
    }
    \label{fig:stage_gain}
\end{figure}

\paragraph{Analysis.}
Fig.~\ref{fig:stage_gain} shows that the effect of human
demonstrations varies across execution stages.
Across tasks, transfer gains are typically strongest in earlier or
intermediate stages and become weaker or less consistent near task
completion.

For pick-and-place and stacking, human demonstrations mainly improve
transport and placement behavior, suggesting that current H2R methods
effectively transfer high-level spatial intent.
In contrast, gains are limited in later stages requiring precise pose
alignment or contact-rich interaction, especially for insertion.

Sequential manipulation shows a similar trend: improvements are often
largest early in execution but diminish over repeated interaction
steps as errors accumulate.

Overall, these results suggest that current H2R methods transfer
coarse spatial behavior more effectively than fine-grained contact
dynamics, with embodiment mismatch becoming most evident in later
stages and long-horizon execution.

\label{sec:supp-stage}

\subsection{Max Human Demos VS Best number of Human Demos}
\begin{table}[H]
\centering
\caption{
Extended benchmark results across all four tasks.
For each task, we report performance under the maximum human-demo
setting (\textbf{Prog.@MaxH}), the best progression score achieved
across the human-demo sweep (\textbf{Best Prog.}), and the
corresponding number of human demonstrations
(\textbf{\# Human}).
}
\label{tab:extended-main}

\footnotesize
\setlength{\tabcolsep}{5pt}
\renewcommand{\arraystretch}{1.05}

\resizebox{\columnwidth}{!}{
\begin{tabular}{lccc|ccc}
\toprule
\multirow{2}{*}{\textbf{Method}}
& \multicolumn{3}{c|}{\textbf{Pick-and-Place}}
& \multicolumn{3}{c}{\textbf{Stacking}} \\
\cmidrule(lr){2-4}
\cmidrule(lr){5-7}
& \textbf{Prog.@MaxH} & \textbf{Best Prog.} & \textbf{\# Human}
& \textbf{Prog.@MaxH} & \textbf{Best Prog.} & \textbf{\# Human} \\
\midrule

Phantom~\cite{lepert2025phantom}
& 76.7 & 76.7 & 100
& 79.2 & 79.2 & 300 \\

Point Policy~\cite{haldar2025point}
& 88.9 & 88.9 & 100
& 88.3 & 93.3 & 100 \\

AMPLIFY~\cite{collins2025amplify}
& 71.1 & 72.2 & 40
& 94.2 & 99.2 & 200 \\

GHOST~\cite{krishna2026ghost}
& 96.7 & 96.7 & 100
& 90.0 & 95.0 & 200 \\

\midrule

\multirow{2}{*}{\textbf{Method}}
& \multicolumn{3}{c|}{\textbf{Insertion}}
& \multicolumn{3}{c}{\textbf{Sequential Manipulation}} \\
\cmidrule(lr){2-4}
\cmidrule(lr){5-7}
& \textbf{Prog.@MaxH} & \textbf{Best Prog.} & \textbf{\# Human}
& \textbf{Prog.@MaxH} & \textbf{Best Prog.} & \textbf{\# Human} \\
\midrule

Phantom~\cite{lepert2025phantom}
& 40.8 & 46.7 & 100
& 36.7 & 36.7 & 300 \\

Point Policy~\cite{haldar2025point}
& 45.0 & 45.0 & 300
& 16.7 & 16.7 & 300 \\

AMPLIFY~\cite{collins2025amplify}
& 41.7 & 41.7 & 100
& 12.2 & 15.6 & 100 \\

GHOST~\cite{krishna2026ghost}
& 65.0 & 65.8 & 100
& 71.1 & 71.1 & 300 \\

\bottomrule
\end{tabular}
}
\end{table}

\subsection{Object Placement Generalization Experiments}
\label{sec:placement_generalization}
To evaluate whether H2R methods can transfer spatial priors from human
demonstrations beyond the robot training distribution, we conduct
object placement generalization experiments on the \textit{Stacking} task.

All methods are trained with \textbf{100 robot demonstrations} and
\textbf{100 human demonstrations}. During robot data collection, the
manipulated object (the \textit{green bowl}) is initialized only on the
\textbf{right side} of the table. In contrast, human demonstrations are
collected with object placements spanning the \textbf{full tabletop
workspace}, covering a broader spatial distribution.

At evaluation time, we intentionally place the object on the
\textbf{left side} of the table, outside the robot training
distribution. This setup isolates each method's ability to leverage
human demonstrations for spatial generalization, measuring whether the
policy can transfer object-placement priors observed in human videos
to unseen robot initializations.

We evaluate four H2R transfer methods:
Phantom~\cite{lepert2025phantom},
AMPLIFY~\cite{collins2025amplify},
Point Policy~\cite{haldar2025point},
and GHOST~\cite{krishna2026ghost}.
We observe that Point Policy achieves the strongest overall performance, obtaining the highest progression score and the highest success rates across later task stages. GHOST achieves the second-best performance, while Phantom and AMPLIFY exhibit substantially larger performance degradation under the distribution shift.

One possible explanation for Point Policy's strong generalization ability is its use of a unified action space for both human and robot demonstrations. By representing demonstrations from both embodiments within the same action space, Point Policy can directly leverage human demonstrations without requiring embodiment-specific action mappings. This unified formulation facilitates knowledge transfer from human demonstrations and may improve robustness to variations in object placement, scene configuration, and task execution.

In contrast, GHOST benefits from human demonstrations primarily through its high-level subgoal policy, while the low-level goal-conditioned policy is trained exclusively on robot demonstrations. As a result, only part of the policy hierarchy directly benefits from human data. Under substantial object placement shifts, the low-level goal-conditioned policy may remain constrained by the diversity of the robot demonstrations and therefore generalize less effectively than Point Policy.

\begin{table}[h]
\centering
\caption{
Object placement generalization evaluation under out-of-distribution
object initialization on the Stacking task. All methods are trained with 100 robot
demonstrations and 100 human demonstrations. Robot demonstrations are
collected with object placement restricted to the right side of the
table, while evaluation is performed with object placement on the left side.
Best per column is \textbf{bolded}; second-best is \underline{underlined}. 
}
\label{tab:placement_generalization_stacking}
\footnotesize
\setlength{\tabcolsep}{4pt}
\renewcommand{\arraystretch}{1.1}
\begin{tabular}{lccccc}
\toprule
\textbf{Method}
& \textbf{Prog.}
& \textbf{SR$_1$}
& \textbf{SR$_2$}
& \textbf{SR$_3$}
& \textbf{SR$_4$} \\
\midrule
Phantom      & 48.3 & 100.0 & 73.3 & 20.0 & 0.0 \\
AMPLIFY      & 41.7 & 100.0 & 60.0 & 6.7  & 0.0 \\
Point Policy & \textbf{78.3} & \textbf{100} & \textbf{100} & \textbf{66.7} & \textbf{46.7} \\
GHOST        & \underline{58.3} & \underline{100} & \underline{86.7} & \underline{30} & \underline{16.7} \\
\bottomrule
\end{tabular}
\end{table}

\subsection{Multi-Seed Evaluation}
\label{sec:multiseed}
To assess whether our benchmark results are sensitive to training
randomness, we additionally evaluate selected experiments across
multiple random seeds. For each selected method--task configuration,
we repeat policy training with different random initializations while
keeping the dataset, hyperparameters, and evaluation protocol fixed.

\begin{figure}[h]
    \centering
    \includegraphics[width=0.92\columnwidth]{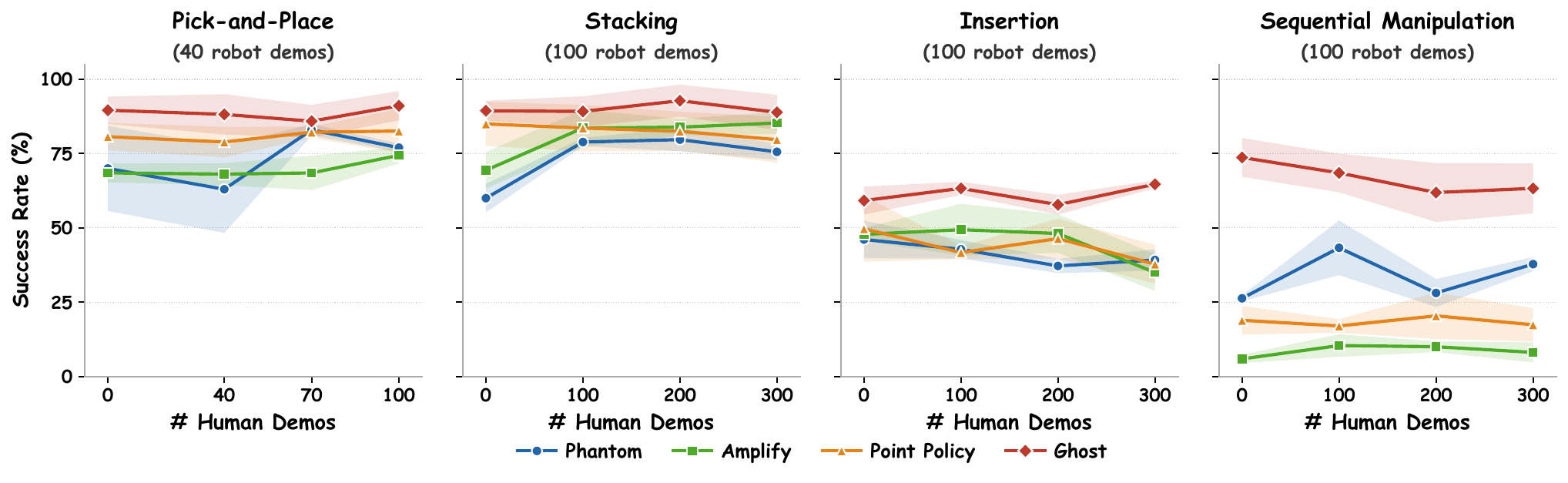}
    \caption{
    Multi-seed human-data scaling performance. Curves show average
    performance across random seeds as the number of human demonstrations
    increases, with shaded regions indicating seed-to-seed variation.
    }
    \label{fig:multiseed_scaling}
\end{figure}

\begin{figure}[h]
    \centering
    \includegraphics[width=\columnwidth]{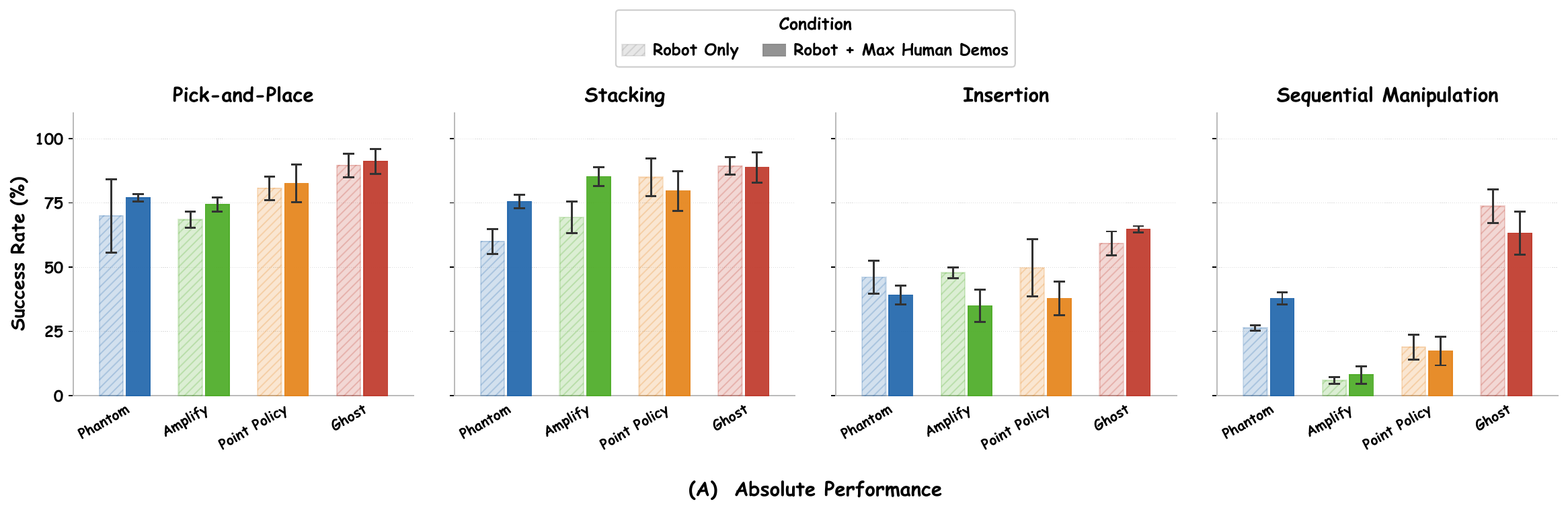}
    \caption{
    Multi-seed absolute performance comparison across H2R methods and tasks.
    Hatched bars indicate robot-only training; solid bars indicate robot + maximum
    human demonstrations. Error bars show standard deviation across random seeds.
    }
    \label{fig:multiseed_absolute}
\end{figure}

We report two complementary multi-seed analyses. First, we summarize
the final performance under the robot + maximum human demonstration
setting, reporting the mean and standard deviation of task progression
(\textbf{Prog.}) and final-stage success rate (\textbf{SR}) across
seeds (Table~\ref{tab:multiseed_main} and Figure~\ref{fig:multiseed_absolute}). Second, we plot human-data scaling curves averaged across
seeds, with error bands indicating seed-to-seed variation
(Figure~\ref{fig:multiseed_scaling}).

These results are intended to verify that the performance trends
reported in the main paper are not dominated by random initialization
or optimization noise. In particular, consistent method ordering and
low variance across seeds would indicate that the observed differences
between H2R methods reflect systematic differences in transfer
behavior rather than stochastic training effects.

\begin{table}[h]
\centering
\caption{
Multi-seed benchmark results under robot + maximum human demonstration
training. We report mean and standard deviation across random seeds
for task progression (\textbf{Prog.}) and final-stage success rate
(\textbf{SR}).
}
\label{tab:multiseed_main}

\scriptsize
\setlength{\tabcolsep}{3.5pt}
\renewcommand{\arraystretch}{0.95}

\resizebox{\columnwidth}{!}{%
\begin{tabular}{lcccccccc}
\toprule
\multirow{2}{*}{\textbf{Method}}
& \multicolumn{2}{c}{\textbf{Pick-and-Place}}
& \multicolumn{2}{c}{\textbf{Stacking}}
& \multicolumn{2}{c}{\textbf{Insertion}}
& \multicolumn{2}{c}{\textbf{Sequential Manipulation}} \\
\cmidrule(lr){2-3}
\cmidrule(lr){4-5}
\cmidrule(lr){6-7}
\cmidrule(lr){8-9}
& \textbf{Prog.} & \textbf{SR}
& \textbf{Prog.} & \textbf{SR}
& \textbf{Prog.} & \textbf{SR}
& \textbf{Prog.} & \textbf{SR} \\
\midrule

Phantom
& 77.0 $\pm$ 1.4 & 56.7 $\pm$ 2.7
& 75.6 $\pm$ 2.7 & 25.6 $\pm$ 6.8
& 39.2 $\pm$ 3.6 & 7.8 $\pm$ 4.2
& 37.8 $\pm$ 2.4 & 13.3 $\pm$ 4.7 \\

AMPLIFY
& 74.4 $\pm$ 2.7 & 58.9 $\pm$ 4.2
& 85.3 $\pm$ 3.7 & 61.1 $\pm$ 9.6
& 35.0 $\pm$ 6.2 & 2.2 $\pm$ 1.6
& 8.1 $\pm$ 3.4 & 0.0 $\pm$ 0.0 \\

Point Policy
& 82.6 $\pm$ 7.4 & 62.2 $\pm$ 11.7
& 79.7 $\pm$ 7.7 & 61.1 $\pm$ 17.1
& 37.8 $\pm$ 6.5 & 6.7 $\pm$ 11.5
& 17.4 $\pm$ 5.6 & 0.0 $\pm$ 0.0 \\

GHOST
& 91.1 $\pm$ 4.9 & 82.2 $\pm$ 6.9
& 88.9 $\pm$ 5.9 & 76.7 $\pm$ 8.8
& 64.7 $\pm$ 1.3 & 34.5 $\pm$ 6.9
& 63.3 $\pm$ 8.4 & 38.9 $\pm$ 15.0 \\

\bottomrule
\end{tabular}
}
\end{table}

\subsection{XSkill (Additional H2R Method Trial)}
We additionally evaluate XSkill~\cite{xu2023xskill}, which trains a skill encoder, a diffusion policy conditioned on the inferred skill representations and a skill alignment transformer. We utilized the left-view observation for skill discovery and all three camera views for skill transferring and composing. Our experiments reveal two failure modes across tasks.

During training, the skill encoder frequently exhibits representation collapse, especially for simpler tasks such as pick-and-place. This degenerate state is validated by the observation that all encoded skill prototypes collapse to a uniform probability distribution, assigning the same score to each prototype at any given timestamp. An analysis of the training loss reveals the mechanics behind this convergence. Specifically, the total loss, defined as$$\mathcal{L} = \lambda_{prototype}\mathcal{L}_{\text{prototype}} + \lambda_{tcn}\mathcal{L}_{\text{tcn}}$$stabilizes at a value of$$\mathcal{L} = 0.5\log K + \log \frac{neg + pos}{pos} \approx 4.57$$when $\lambda_{prototype}=0.5$, $\lambda_{tcn}=1$, $K=32$ prototypes are used, $neg = 16$ negative clips and $pos = 1$ positive clip are sampled (utilizing the real-world hyperparameters settings reported in the original paper). This stabilization point corresponds to the theoretical minimum of the loss under a fully collapsed representation. When the encoder maps all video clips to a singular point in the embedding space, the Sinkhorn-Knopp algorithm assigns uniform target probabilities $\frac{1}{K}$ to each prototype. Consequently, the predicted probabilities also become uniform at $\frac{1}{K}$, yielding $\mathcal{L}_{\text{prototype}} = \log K$, while the time-contrastive loss $\mathcal{L}_{\text{tcn}}$ simultaneously reduces to $\log \frac{neg + pos}{pos}$. At this theoretical minimum, the gradients vanish and training stalls. Without meaningful variance in the encoded skill representations, the skill-conditioned diffusion policy subsequently degrades into an unconditioned diffusion policy. We hypothesize that this representation collapse is because simpler tasks feature fewer distinct manipulation phases and therefore lack the visually separable states necessary to drive prototype specialization.

\begin{figure}[htbp]
    \centering
 
    \begin{subfigure}[t]{\columnwidth}
        \centering
        \includegraphics[width=0.92\linewidth]{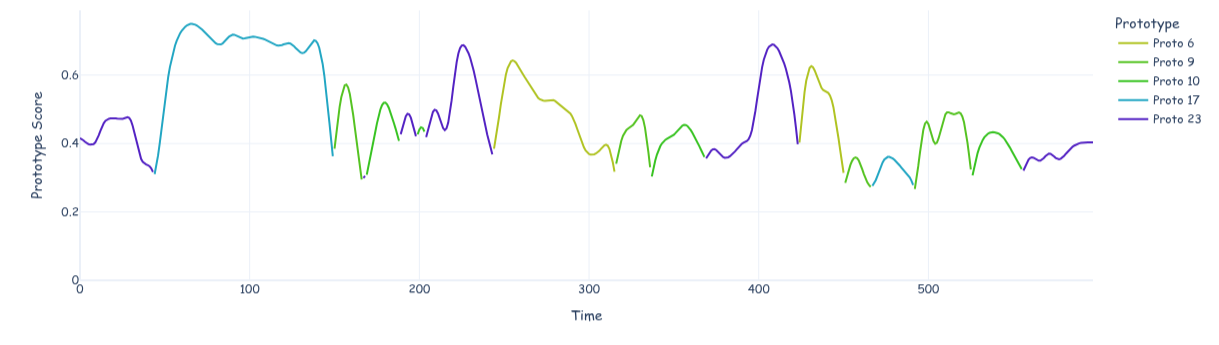}
        \label{fig:XSkill_human}
    \end{subfigure}
    \begin{subfigure}[t]{\columnwidth}
        \centering
        \includegraphics[width=0.92\linewidth]{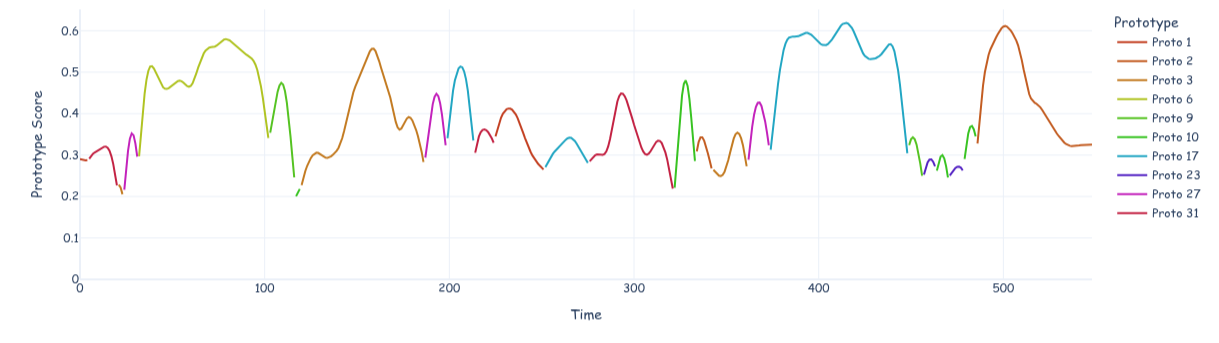}
        \label{fig:XSkill_robot}
    \end{subfigure}
 
    \caption{
    Encoded prototype scores of random selected demonstration from sequential manipulation task. Different colors represent different prototypes. For visual clarity, only the prototype with the highest score at each time step is plotted.
    \textbf{Up}: human demo.
    \textbf{Down}: robot demo.
    }
    \label{fig:XSkill_encoded_protos}
\end{figure}

For more complex tasks, the skill encoder does converge, but performance drops substantially compared to simply training an unconditioned diffusion policy using robot demonstrations, as shown in Tab.~\ref{tab:xskill}. To probe whether the converged encoder captures semantically meaningful skills, we recorded a human demonstration of the sequential manipulation task with a reordered picking sequence (original: blue $\rightarrow$ orange $\rightarrow$ yellow; new: orange $\rightarrow$ blue $\rightarrow$ yellow) and used it as the prompt video at inference time. Across 10 rollouts, the robot consistently reached for the blue toy first without hesitation, indicating that it ignored the prompt entirely. In Fig.~\ref{fig:XSkill_encoded_protos}, We also visualize the encoded prototype scores for random selected human and robot demonstration in the sequential manipulation task ; no cross-embodiment alignment is apparent. We hypothesize that the skill encoder fails to learn semantically grounded skill representations, and instead captures low-level visual statistics or temporal progress that happen to be stable under augmentation. The uninformative skill representations introduce noise to the diffusion policy and degrade policy's performance relative to the unconditioned baseline.

We identify the primary failure mode as insufficient diversity in skill ordering across demonstrations. XSkill's cross-embodiment alignment relies on the same skills appearing in varying sequential contexts so that the encoder is pressured to represent skill identity. In our tasks, however, all human and robot demonstrations follow a fixed object-picking order, even for sequential manipulation task. This contrasts with the original XSkill real-world experiments, where the three kitchen sub-tasks (Draw, Light, Oven) are executed in different orders across demonstrations, providing the combinatorial diversity necessary to disentangle skill identity from temporal context.
\begin{table}[h]
\centering
\caption{
Simulation results on three additional tasks. Diffusion Policy is
trained with 100 robot demonstrations. XSkill is trained with 100 human
demonstrations and 100 robot demonstrations.
}
\label{tab:xskill}

\scriptsize
\setlength{\tabcolsep}{4pt}
\renewcommand{\arraystretch}{0.95}

\begin{tabular*}{\columnwidth}{@{\extracolsep{\fill}}lcccccc@{}}
\toprule

\multirow{2}{*}{\textbf{Method}}
& \multicolumn{2}{c}{\textbf{Stacking}}
& \multicolumn{2}{c}{\textbf{Insertion}}
& \multicolumn{2}{c}{\textbf{Sequential Manipulation}} \\

\cmidrule(lr){2-3}
\cmidrule(lr){4-5}
\cmidrule(lr){6-7}

& \textbf{Prog.} & \textbf{SR}
& \textbf{Prog.} & \textbf{SR}
& \textbf{Prog.} & \textbf{SR} \\

\midrule

Diffusion Policy~\cite{chi2024diffusionpolicy}
& 95.0 & 80.0
& 41.7 & 0.0
& 82.2 & 63.3 \\

XSkill~\cite{xu2023xskill}
& 28.3 & 0.0
& 10.0 & 0.0
& 10.0 & 0.0 \\

\bottomrule
\end{tabular*}
\end{table}
 
\newpage
\section{Failure Cases and Analyses}
\label{sec:failure_case_analysis}

\subsection{Method-Specific Analysis}

\subsubsection{Phantom Failure Cases}
\label{sec:phantom_failure_cases}
As shown in the main benchmark results, Phantom performs
competitively on coarse manipulation tasks but remains more
challenging on tasks requiring precise contact-rich interaction such
as stacking and insertion. To better understand this behavior, we
analyze the generated robot-inpainted demonstrations used as Phantom’s
training supervision.

\begin{figure}[H]
    \centering

    \begin{subfigure}[t]{0.98\columnwidth}
        \centering
        \includegraphics[width=\linewidth]{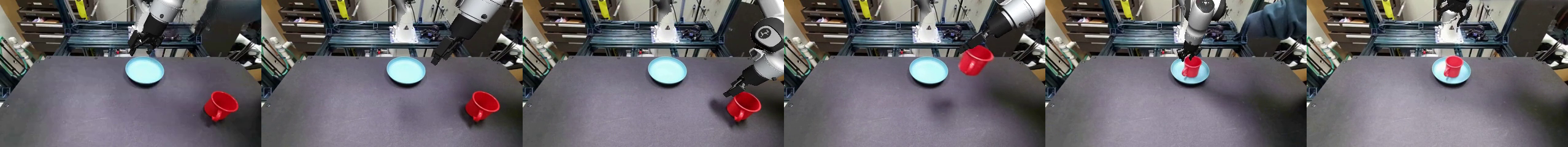}
        \caption{Pick-and-Place}
    \end{subfigure}

    \vspace{0.6em}

    \begin{subfigure}[t]{0.98\columnwidth}
        \centering
        \includegraphics[width=\linewidth]{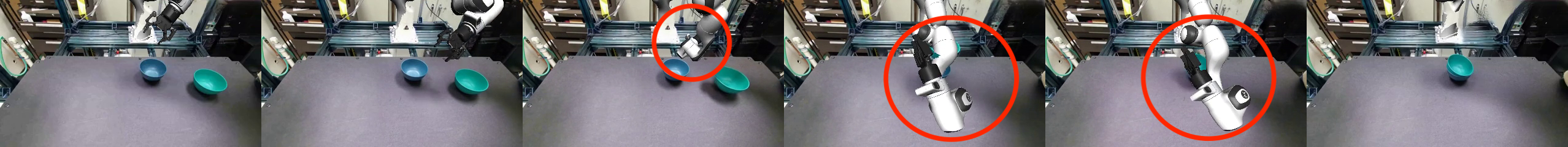}
        \caption{Stacking}
    \end{subfigure}

    \vspace{0.6em}

    \begin{subfigure}[t]{0.98\columnwidth}
        \centering
        \includegraphics[width=\linewidth]{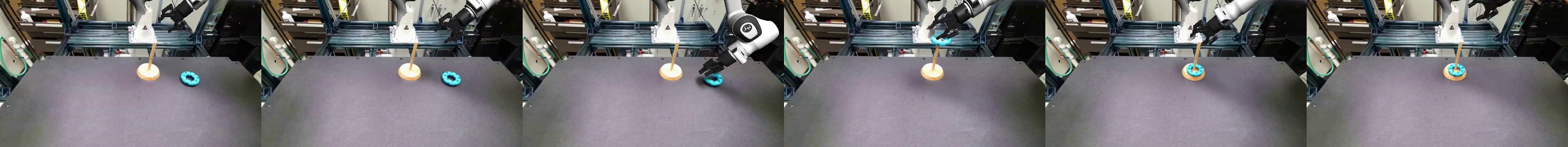}
        \caption{Insertion}
    \end{subfigure}

    \vspace{0.6em}

    \begin{subfigure}[t]{0.98\columnwidth}
        \centering
        \includegraphics[width=\linewidth]{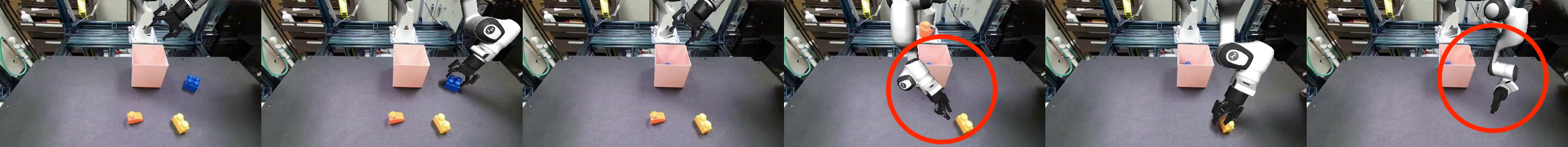}
        \caption{Sequential Manipulation}
    \end{subfigure}

    \caption{
    Representative robot-inpainting results generated by
Phantom~\cite{lepert2025phantom} across the four H2RBench tasks.
    }
    \label{fig:phantom_failure_cases}
\end{figure}

Figure~\ref{fig:phantom_failure_cases} visualizes representative
robot-inpainting results across the four H2RBench tasks. Phantom
bridges the embodiment gap by retargeting human hand motion into robot
end-effector trajectories and rendering corresponding robot
demonstrations through inverse kinematics. While this produces
visually plausible robot behavior, we observe two recurring sources of
error.

First, the generated robot images occasionally exhibit visual
artifacts or imperfect alignment between the rendered robot and the
scene, introducing noise in the visual supervision signal.

More importantly, the retargeted robot actions are often not
geometrically precise with respect to the manipulated objects. Across
tasks, the generated end-effector trajectory can deviate slightly from
the intended grasp pose, contact location, or placement configuration.
These errors are typically small at individual timesteps but become
particularly noticeable in precise manipulation settings where task
success depends strongly on accurate object contact and relative pose.

Together, these observations suggest that Phantom’s limitations arise
not only from image-generation quality, but also from inaccuracies in
human-to-robot motion retargeting. Since the policy is trained on
these generated robot demonstrations, imprecision in the retargeted
actions can directly propagate into the learned behavior, making
fine-grained manipulation tasks more difficult to transfer reliably.

\subsubsection{Amplify Failure Cases}
\label{sec:amplify_failure_cases}

To better understand the task-dependent performance of
AMPLIFY~\cite{collins2025amplify}, we further analyze the learned
motion latent space using two complementary metrics: (1) the
2-Wasserstein (Fréchet) distance between human and robot latent
distributions, and (2) t-SNE visualization of latent embeddings.

Following recent analysis in~\cite{punamiya2025egobridge}, we compute
the 2-Wasserstein distance ($W_2$) between human and robot latent
features using a Gaussian approximation in PCA-reduced latent space.
This metric captures both centroid shift and covariance mismatch
between the two distributions, and provides a quantitative measure of
cross-embodiment alignment.

Table~\ref{tab:amplify_w2_distance} summarizes the $W_2$ distance
across tasks under different human demonstration scales. We observe
that \textbf{Stacking}, the task where AMPLIFY achieves the strongest
performance improvement in simulation, exhibits a monotonic decrease
in $W_2$ distance as more human demonstrations are added. This
suggests that the motion tokenizer learns increasingly aligned latent
representations between human and robot trajectories, improving
cross-embodiment transfer.

In contrast, \textbf{Insertion} shows a monotonic increase in $W_2$
distance with additional human demonstrations, indicating growing
distributional mismatch between human and robot motions. We hypothesize
that this is caused by the fundamentally different interaction
dynamics between human fingertip insertion and robot gripper-based peg
alignment, which makes latent alignment more difficult and limits
transfer performance.

\begin{figure}[H]
    \centering

    \begin{subfigure}[t]{0.47\columnwidth}
        \centering
        \includegraphics[width=\linewidth]{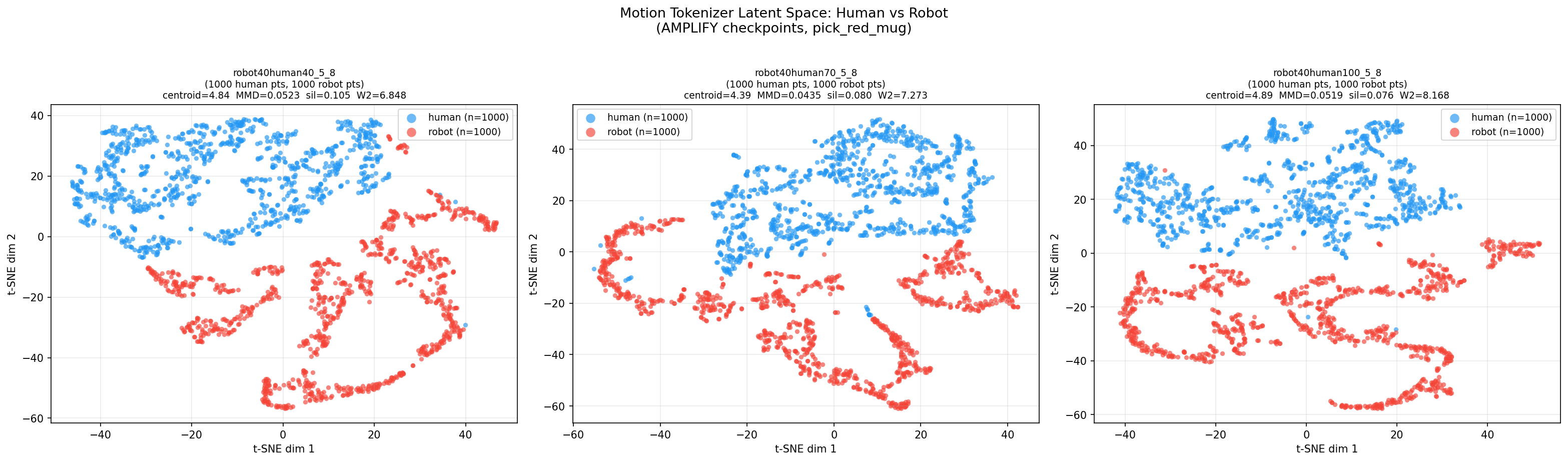}
        \caption{Pick-and-Place}
    \end{subfigure}
    \hfill
    \begin{subfigure}[t]{0.47\columnwidth}
        \centering
        \includegraphics[width=\linewidth]{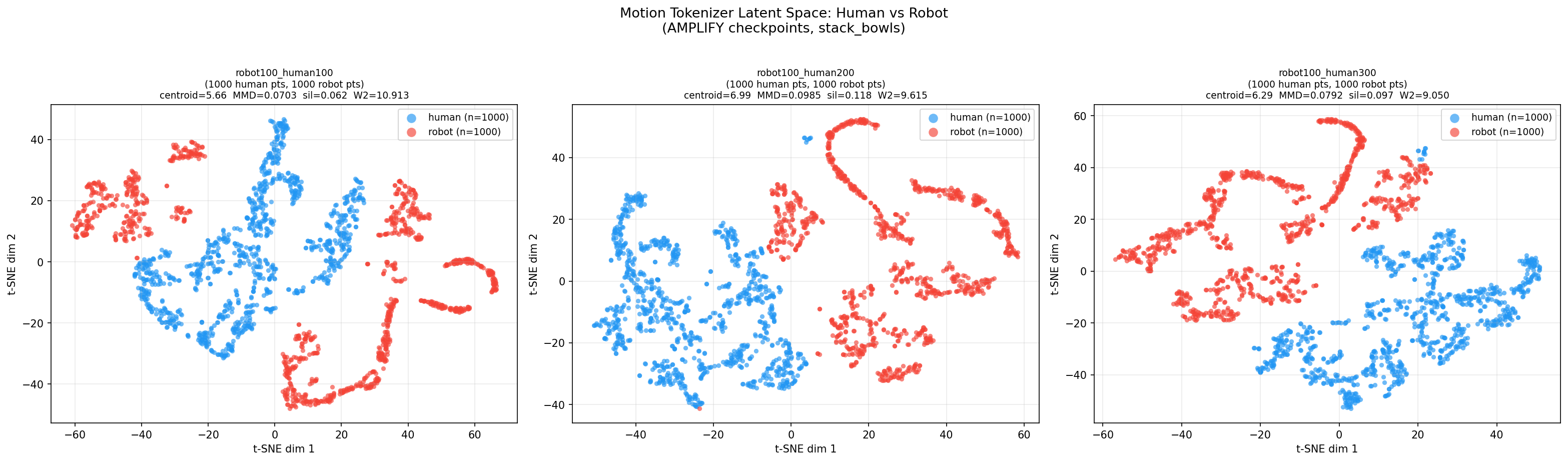}
        \caption{Stacking}
    \end{subfigure}

    \vspace{0.4em}

    \begin{subfigure}[t]{0.47\columnwidth}
        \centering
        \includegraphics[width=\linewidth]{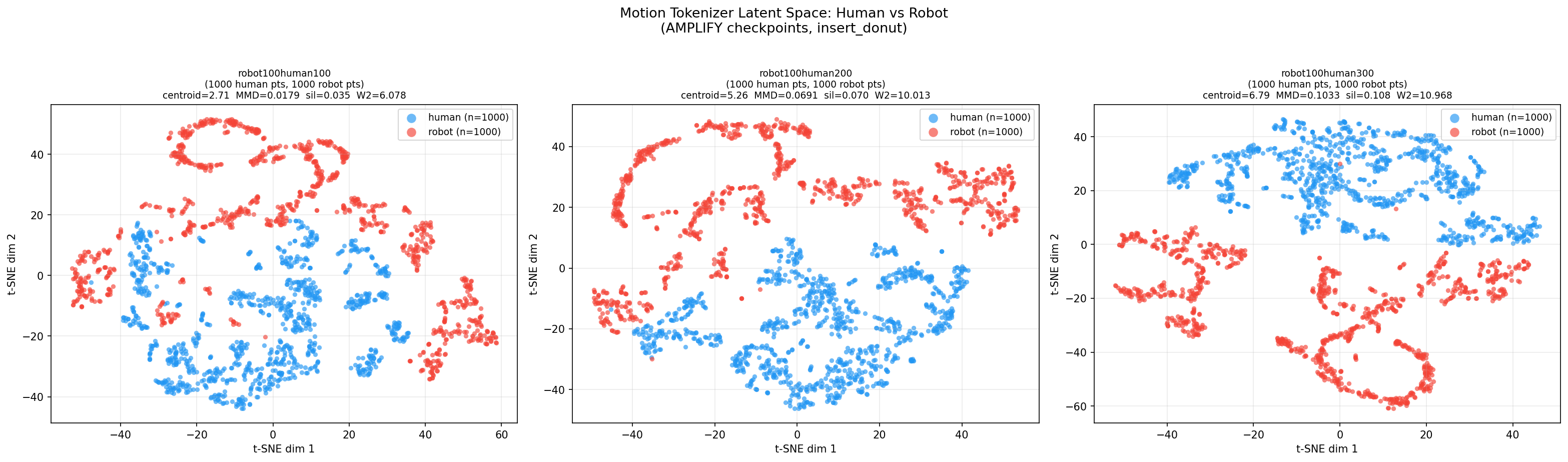}
        \caption{Insertion}
    \end{subfigure}
    \hfill
    \begin{subfigure}[t]{0.47\columnwidth}
        \centering
        \includegraphics[width=\linewidth]{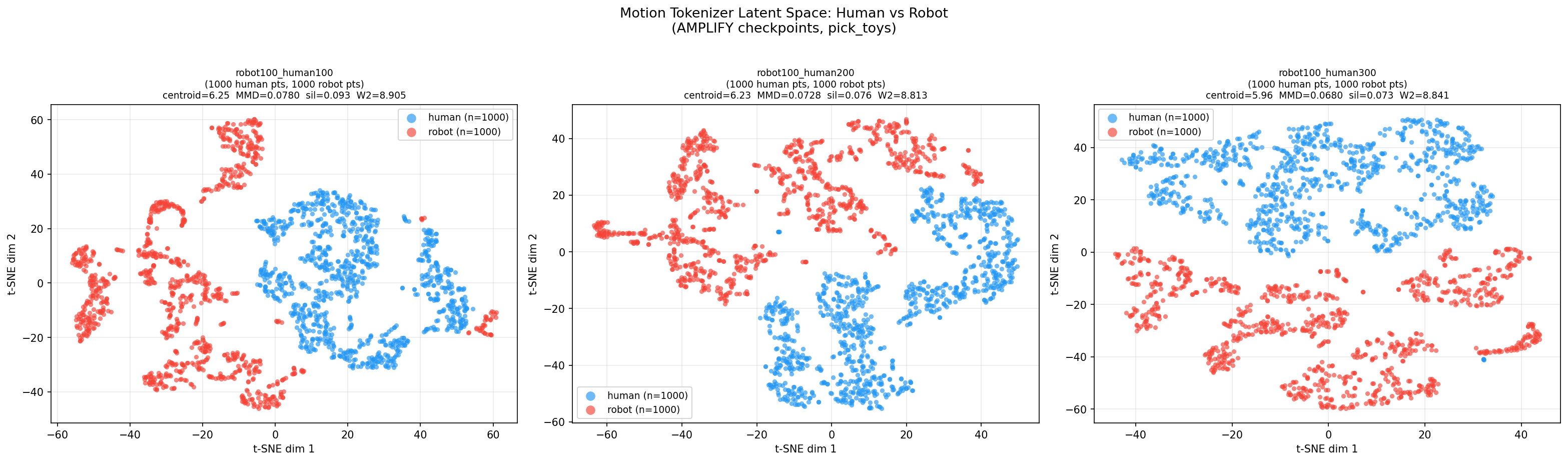}
        \caption{Sequential Manipulation}
    \end{subfigure}

    \caption{
    t-SNE visualization of AMPLIFY motion latent embeddings for human
    and robot demonstrations across four H2RBench tasks. Human and
    robot trajectories form partially separated clusters with limited
    overlap, indicating incomplete alignment of the learned
    cross-embodiment latent representation. The degree of overlap
    varies across tasks and correlates with downstream H2R transfer
    performance.
    }
    \label{fig:amplify_tsne}
\end{figure}

Figure~\ref{fig:amplify_tsne} further visualizes the latent embeddings
for human and robot demonstrations using t-SNE. Across all four tasks,
human and robot trajectories form partially separated clusters with
limited overlap. This suggests that AMPLIFY does not fully learn a
shared cross-embodiment latent skill space. Instead, the learned
representation remains partially domain-specific, which may explain
why performance gains from human demonstrations are task-dependent and
why transfer remains challenging in some manipulation settings.

Together, these results suggest that latent distribution alignment is
closely correlated with AMPLIFY's H2R transfer performance: tasks with
stronger human--robot latent alignment tend to benefit more from human
demonstrations, while tasks with persistent distributional separation
remain more difficult to transfer.

\begin{table}[H]
\centering
\caption{
2-Wasserstein ($W_2$) distance between human and robot latent feature
distributions for AMPLIFY under different human demonstration scales.
Lower values indicate stronger alignment between human and robot latent
representations.
}
\label{tab:amplify_w2_distance}

\footnotesize
\setlength{\tabcolsep}{7pt}
\renewcommand{\arraystretch}{1.05}

\begin{tabular}{lccc}
\toprule
\textbf{Task}
& \textbf{Low Human}
& \textbf{Mid Human}
& \textbf{High Human} \\
\midrule

Stacking
& 10.91
& 9.62
& 9.05 \\

Pick-and-Place
& 6.85
& 7.27
& 8.17 \\

Insertion
& 6.08
& 10.01
& 10.97 \\

Sequential Manipulation
& 8.91
& 8.81
& 8.84 \\

\bottomrule
\end{tabular}
\end{table}

\subsubsection{Point Policy Failure Cases}

As shown in the main benchmark results, Point Policy~\cite{haldar2025point} performs well across a variety of manipulation tasks, but its performance drops significantly on long-horizon pick-and-place tasks. To better understand this limitation, we analyze the visual correspondence and tracking pipeline used to construct the policy state representation.

Point Policy uses DIFT~\cite{tang2023emergentcorrespondenceimagediffusion} to transfer task-relevant object keypoints from a manually annotated reference image to the first frame of an observed trajectory. These keypoints are then used to initialize Co-Tracker~\cite{karaev2024cotrackerbettertrack}, which tracks them throughout the remainder of the trajectory. The resulting point trajectories serve as the primary state representation provided to the policy.

\begin{figure}[H]
    \centering

    \begin{subfigure}[t]{0.98\columnwidth}
        \centering
        \includegraphics[width=\linewidth]{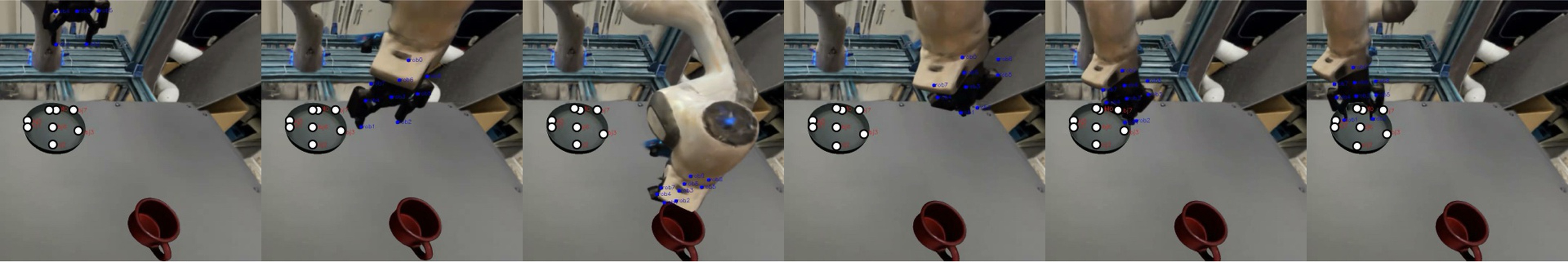}
        \caption{Pick-and-Place}
    \end{subfigure}

    \vspace{0.6em}

    \begin{subfigure}[t]{0.98\columnwidth}
        \centering
        \includegraphics[width=\linewidth]{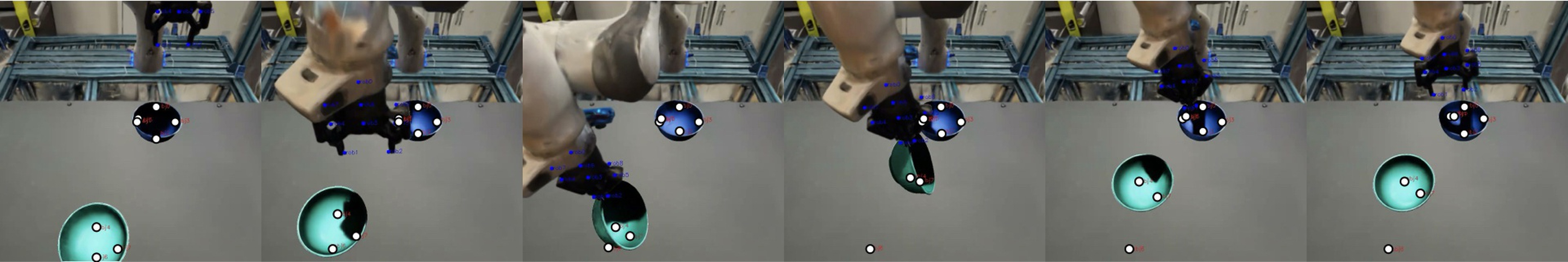}
        \caption{Stacking}
    \end{subfigure}

    \vspace{0.6em}

    \begin{subfigure}[t]{0.98\columnwidth}
        \centering
        \includegraphics[width=\linewidth]{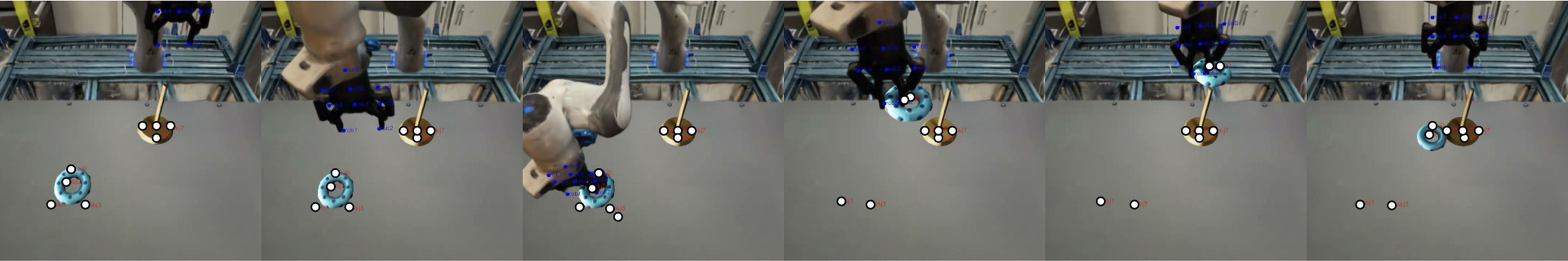}
        \caption{Insertion}
    \end{subfigure}

    \vspace{0.6em}

    \begin{subfigure}[t]{0.98\columnwidth}
        \centering
        \includegraphics[width=\linewidth]{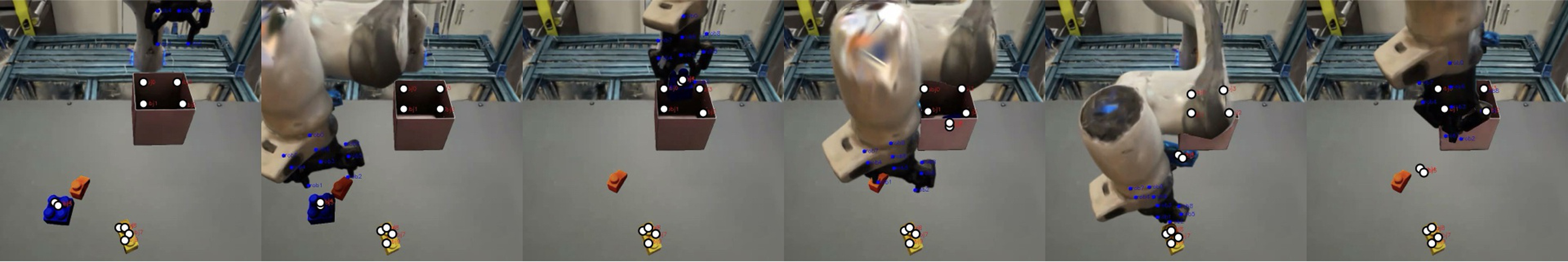}
        \caption{Sequential Manipulation}
    \end{subfigure}

    \caption{
    Representative keypoint tracking visualizations of Point Policy~\cite{haldar2025point}
    during evaluation rollouts across the four H2RBench tasks. All shown rollouts are failure cases.
    White markers denote task-relevant object keypoints, while blue markers denote robot keypoints.
    }
    \label{fig:point_policy_tracking_visualization}
\end{figure}

Figure~\ref{fig:point_policy_tracking_visualization} shows representative tracked keypoints during evaluation rollouts. We observe two common failure modes. First, DIFT can occasionally produce incorrect correspondences due to appearance changes, lighting variations, occlusions, or visually ambiguous object regions. Since these correspondences determine the initial points used by Co-Tracker, errors at this stage can cause the tracker to follow semantically incorrect regions throughout the trajectory. Second, Co-Tracker can struggle when tracked points become partially or fully occluded during task execution. This issue frequently arises in scenarios involving object handoffs, container interactions, or sequential pick-and-place operations, where task-relevant objects may become temporarily or permanently invisible to the camera. In these cases, tracked points can drift, become unstable, or disappear altogether.

These tracking failures are particularly problematic in long-horizon pick-and-place tasks. After an object is placed into a container, it often becomes fully occluded from the camera's view while Co-Tracker continues tracking its associated keypoints. Since the object is no longer visible, the tracker may drift to unrelated image regions and produce meaningless trajectories. These erroneous trajectories remain part of the policy state representation, introducing noise that accumulates over time. As more objects are manipulated and become occluded, the state representation becomes increasingly unreliable, making it harder for the policy to reason about the remaining objects and execute subsequent placements. This failure mode helps explain the near-zero success rates observed when placing later objects, even when the first placement is completed successfully.

Although the policy is trained on demonstrations that naturally contain imperfect correspondences and tracking noise, its performance still depends heavily on the quality of the resulting visual state representation. Errors from either DIFT or Co-Tracker directly affect the point trajectories provided to the policy and can therefore propagate to downstream decisions. This sensitivity is especially evident in tasks requiring precise object contact, accurate placement, or fine-grained spatial reasoning, where even small perception errors can lead to cascading failures.

Overall, these results suggest that Point Policy is highly dependent on the reliability of the underlying correspondence and tracking modules. Improving correspondence quality, handling object occlusions more effectively, and developing more robust long-horizon tracking methods remain important directions for future work.

\subsubsection{GHOST Failure Cases}

As shown in the main benchmark results, GHOST~\cite{krishna2026ghost} performs competitively across a range of manipulation tasks and demonstrates strong task-level reasoning capabilities. Nevertheless, failures still occur, particularly in tasks requiring precise object contact, alignment, or placement. To better understand these limitations, we analyze two common sources of error in the GHOST pipeline.

\begin{figure}[H]
    \centering

    \begin{subfigure}[t]{0.98\columnwidth}
        \centering
        \includegraphics[width=\linewidth]{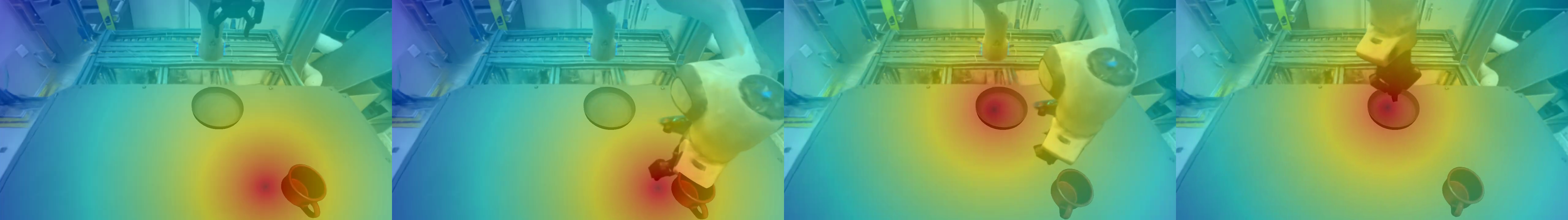}
        \caption{Pick-and-Place}
    \end{subfigure}

    \vspace{0.6em}

    \begin{subfigure}[t]{0.98\columnwidth}
        \centering
        \includegraphics[width=\linewidth]{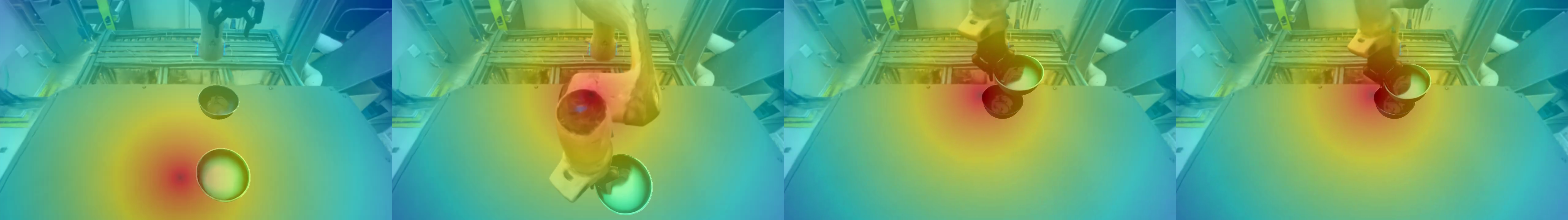}
        \caption{Stacking}
    \end{subfigure}

    \vspace{0.6em}

    \begin{subfigure}[t]{0.98\columnwidth}
        \centering
        \includegraphics[width=\linewidth]{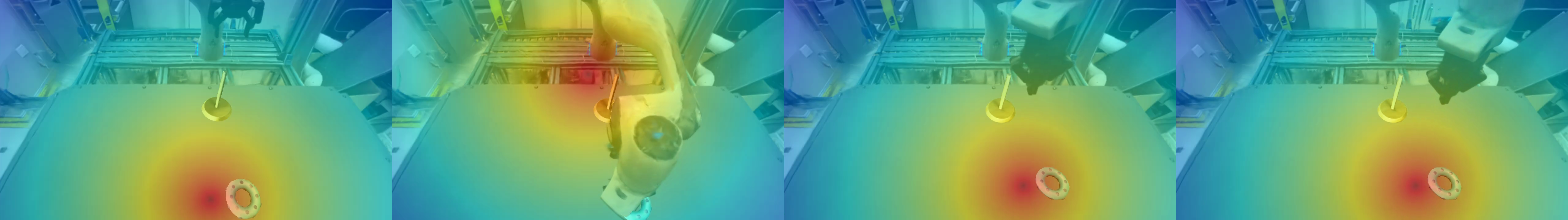}
        \caption{Insertion}
    \end{subfigure}

    \vspace{0.6em}

    \begin{subfigure}[t]{0.98\columnwidth}
        \centering
        \includegraphics[width=\linewidth]{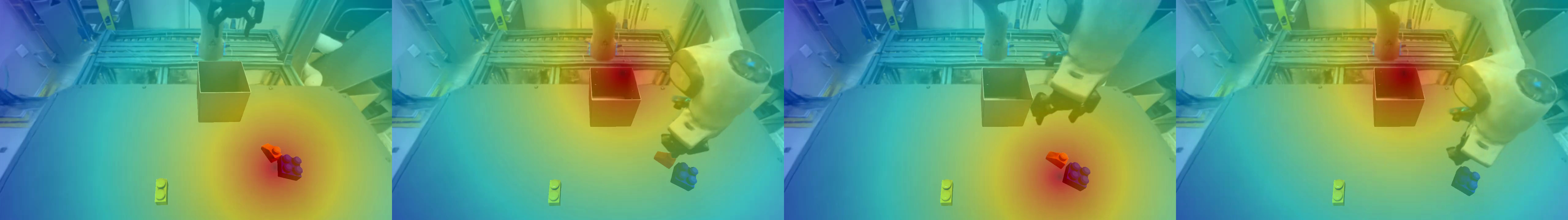}
        \caption{Sequential Manipulation}
    \end{subfigure}

    \caption{
Representative high-level prediction visualizations of GHOST~\cite{krishna2026ghost} during evaluation rollouts across the four H2RBench tasks. The heatmaps are generated by projecting the high-level 3D goal prediction onto the image plane. All shown rollouts correspond to failure cases. Additional rollout videos are available on the project website.
}

    \label{fig:ghost_tracking_visualization}
\end{figure}

First, many failures appear to originate from limitations in the low-level goal-conditioned policy rather than the high-level goal prediction module. During evaluation, we frequently observe cases where the predicted high-level subgoals are semantically correct and correspond to the appropriate object interaction strategy. However, the low-level policy occasionally predicts actions that follow the intended subgoal but are not sufficiently precise to successfully complete the manipulation, resulting in small spatial deviations in the executed end-effector motion. While such deviations are often inconsequential for coarse manipulation tasks, they become increasingly problematic in tasks requiring precise grasping, insertion, alignment, or object placement. As shown in Figure~\ref{fig:ghost_tracking_visualization}, in those failed rollouts, the robot often approaches the correct target and follows the predicted high-level goal, yet ultimately fails because the end-effector misses the intended contact location by a small margin. These observations suggest that the primary bottleneck of GHOST in challenging manipulation scenarios is the precision of the low-level goal-conditioned policy rather than high-level subgoal prediction.

A second source of error stems from imperfect 3D hand pose estimation in human demonstrations. Similar to prior H2R methods that rely on vision-based hand tracking, GHOST depends on estimated 3D hand states to infer human-object interactions and extract task-relevant supervision. In practice, 3D hand pose estimation is inherently noisy and can become unreliable under occlusion, motion blur, self-contact, or challenging viewpoints, leading to inaccurate estimates of hand configuration and object interaction. Since the high-level supervision used by GHOST is derived directly from these estimated hand trajectories, errors in hand pose estimation can propagate into the extracted 3D goals and subgoals used for training. Consequently, inaccuracies in hand tracking may degrade the quality of supervision available to the high-level goal prediction module and contribute to downstream task failures.

In addition, GHOST extracts subgoals using heuristic rules based on estimated gripper open and close events. While this strategy is effective across all tasks evaluated in H2RBench, such heuristics may not always capture the underlying task structure and remain sensitive to errors in hand-state estimation. This limitation is particularly relevant for tasks involving prolonged object contact or continuous manipulation, where meaningful task transitions may not coincide with discrete grasping events. Consequently, inaccuracies in hand-state estimation or subgoal extraction can occasionally lead to suboptimal supervision and contribute to downstream execution failures.

Therefore, these observations suggest that GHOST's remaining failures are primarily associated with limitations in the low-level goal-conditioned policy and imperfect 3D hand pose estimation. Addressing these limitations through more accurate low-level goal-conditioned policies, improved 3D hand pose estimation methods, and more robust goal and subgoal extraction techniques represents a promising direction for future work.

\section{Real-World Experiments}
To evaluate whether H2RBench simulation performance is predictive of
real-world deployment, we benchmark all evaluated H2R methods on the
physical robot under both robot-only and maximum-human-demonstration
training settings. In addition to the main-paper results, we provide
further analysis of Sim-vs.-Real correspondence from three
complementary perspectives: overall benchmark-level correlation,
per-task consistency, and sensitivity to individual methods and tasks.

\subsection{Overall Sim-vs.-Real Performance}
Table~\ref{tab:simreal_overall} summarizes overall Sim-vs.-Real
fidelity across all four tasks and four evaluated H2R methods.
Pearson $r$ and Spearman $\rho$ are computed over all method--task
pairs, while MMRV is computed per task and averaged across tasks.

Overall, simulation performance remains strongly predictive of
real-world performance under both training settings. In the robot-only
setting, we observe near-perfect linear correlation
($r=0.971$) together with strong rank consistency
($\rho=0.940$). Under maximum human supervision, correlation remains
high ($r=0.894$; $\rho=0.851$), although with slightly larger
variation in relative method ordering.

\begin{figure}[H]
    \centering

    \begin{subfigure}[t]{0.48\columnwidth}
        \centering
        \includegraphics[width=\linewidth]{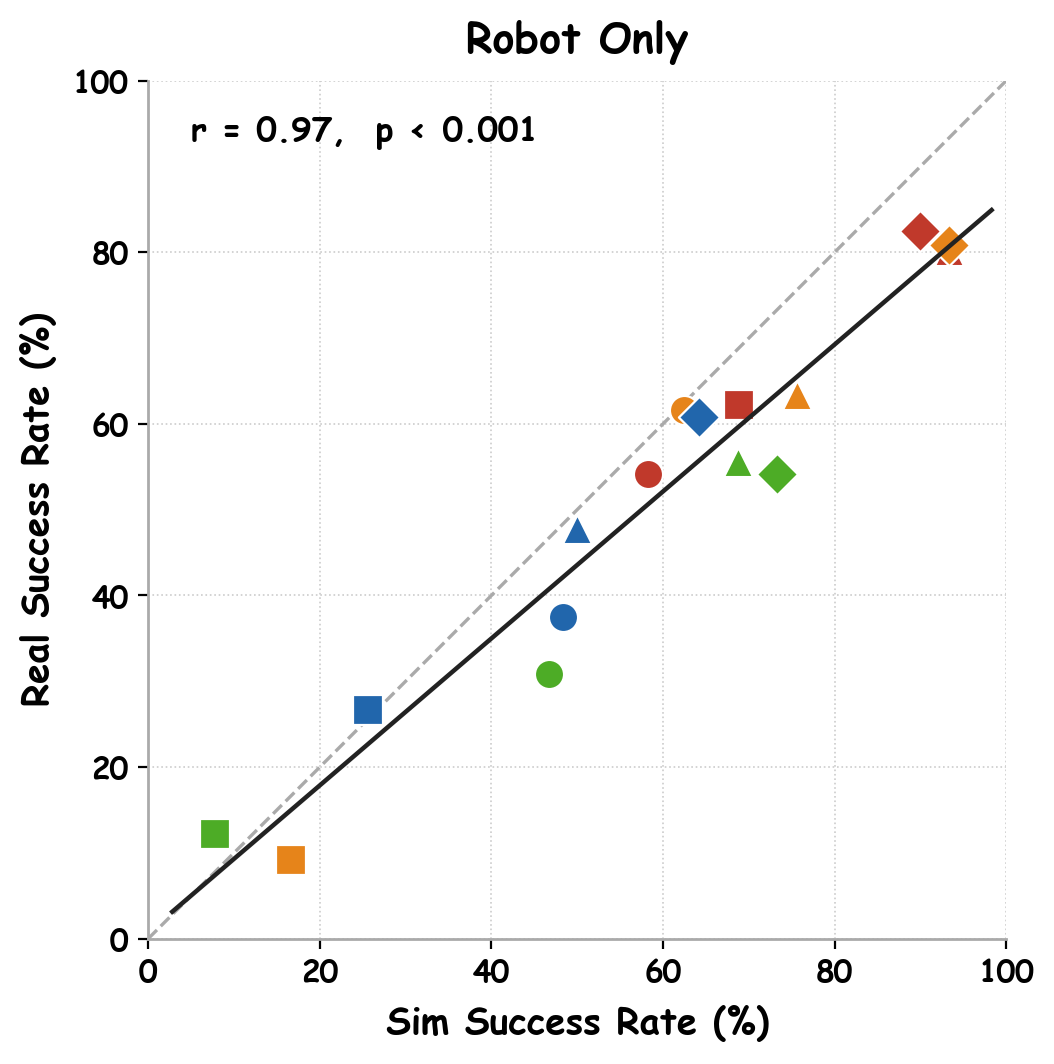}
        \caption{Robot-only}
        \label{fig:simreal_robotonly}
    \end{subfigure}
    \hfill
    \begin{subfigure}[t]{0.48\columnwidth}
        \centering
        \includegraphics[width=\linewidth]{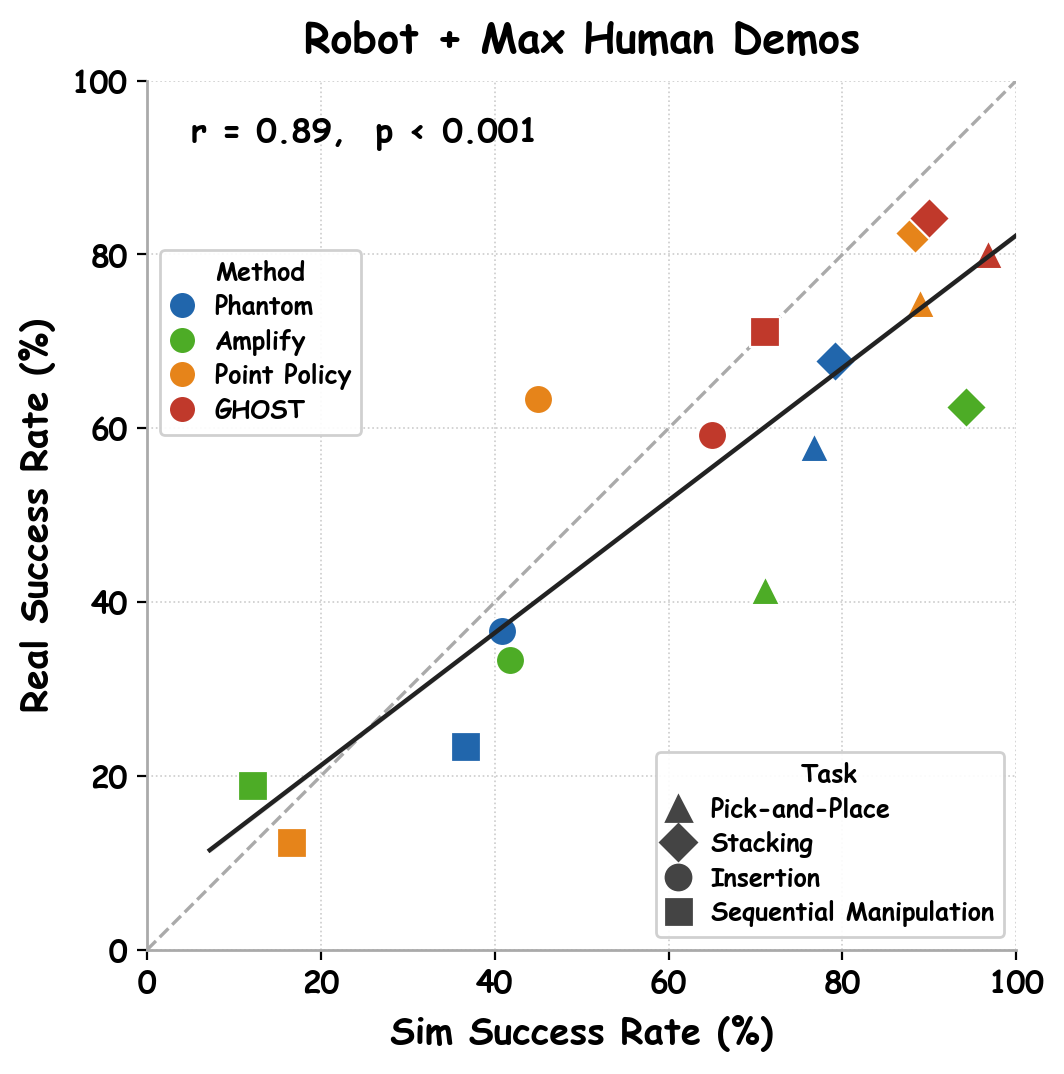}
        \caption{Robot + Max Human}
        \label{fig:simreal_maxhuman}
    \end{subfigure}

    \caption{
    Sim-vs.-Real performance correlation under robot-only (left) and
robot + maximum human demonstration (right) settings. Each point
corresponds to one method--task pair. Simulated performance is broadly
predictive of real-world performance across both training conditions.
    }
    \label{fig:per_task_simreal}
\end{figure}

Figure~\ref{fig:per_task_simreal} visualizes this relationship across
all method--task pairs. Across both settings, higher simulated success
rates generally correspond to higher real-world success rates,
indicating that H2RBench captures the relative performance trends of
H2R methods reliably before deployment on physical hardware.
\begin{table}[H]
\centering
\caption{
Overall Sim-vs.-Real fidelity metrics across four tasks and four H2R
methods. Pearson $r$ and Spearman $\rho$ are computed over all
method--task pairs ($N=16$). MMRV is computed per task using
normalized success rates in $[0,1]$ and averaged across tasks.
Lower MMRV indicates stronger agreement in relative method ordering.
}
\label{tab:simreal_overall}

\small
\setlength{\tabcolsep}{6pt}
\renewcommand{\arraystretch}{1.05}

\begin{tabular}{lccc}
\toprule
\textbf{Condition}
& \textbf{Pearson $r$ $\uparrow$}
& \textbf{Spearman $\rho$ $\uparrow$}
& \textbf{MMRV $\downarrow$} \\
\midrule
Robot Only
& 0.971
& 0.940
& 0.014 \\

Robot + Max Human
& 0.894
& 0.851
& 0.060 \\
\bottomrule
\end{tabular}
\end{table}

\subsection{Per-Task Sim vs. Real Correlation}

\begin{table}[H]
\centering
\caption{
Per-task Mean Maximum Rank Violation (MMRV) under robot-only and
maximum-human-demonstration settings.
}
\label{tab:simreal_task_mmrv}

\small
\setlength{\tabcolsep}{6pt}
\renewcommand{\arraystretch}{1.05}

\begin{tabular}{lcc}
\toprule
\textbf{Task}
& \textbf{Robot Only $\downarrow$}
& \textbf{Robot + Max Human $\downarrow$} \\
\midrule
Pick-and-Place
& 0.000
& 0.000 \\

Stacking
& 0.042
& 0.171 \\

Insertion
& 0.000
& 0.037 \\

Sequential Manipulation
& 0.015
& 0.033 \\
\midrule
\textbf{Mean}
& \textbf{0.014}
& \textbf{0.060} \\
\bottomrule
\end{tabular}
\end{table}

\begin{figure}[H]
    \centering
    \includegraphics[width=\columnwidth]{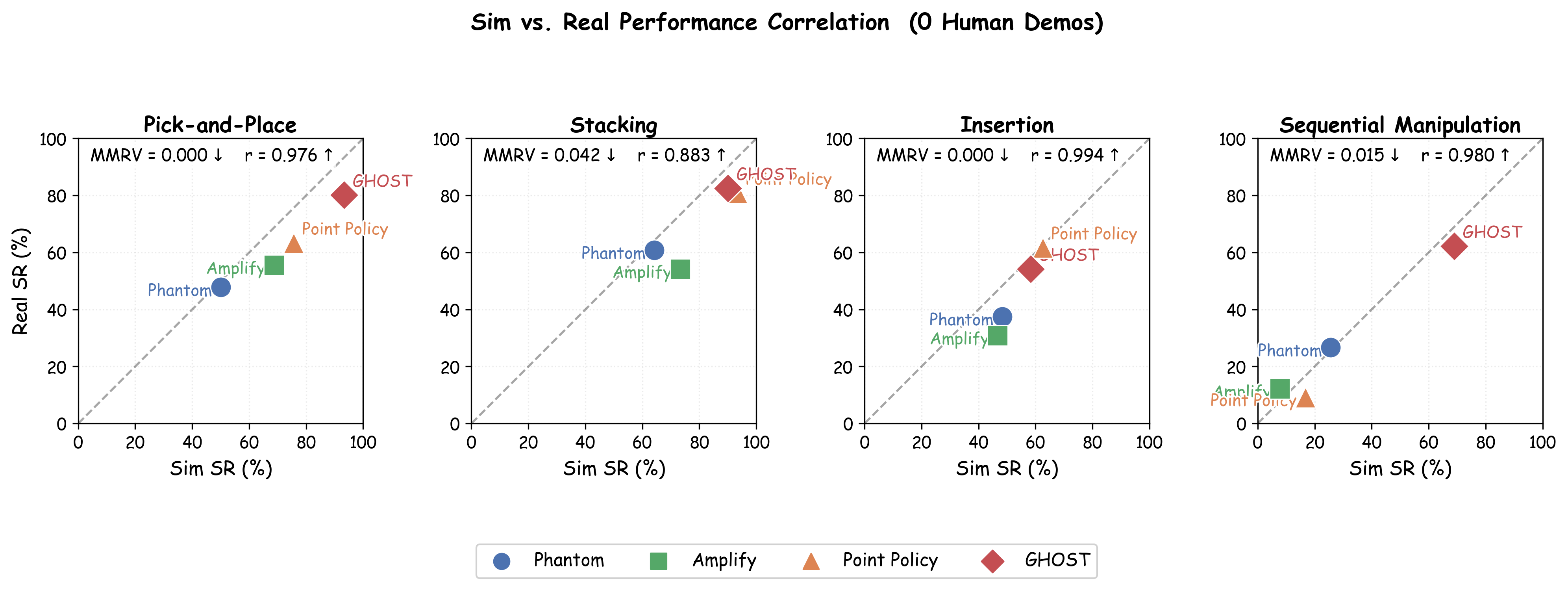}

    \vspace{4pt}

    \includegraphics[width=\columnwidth]{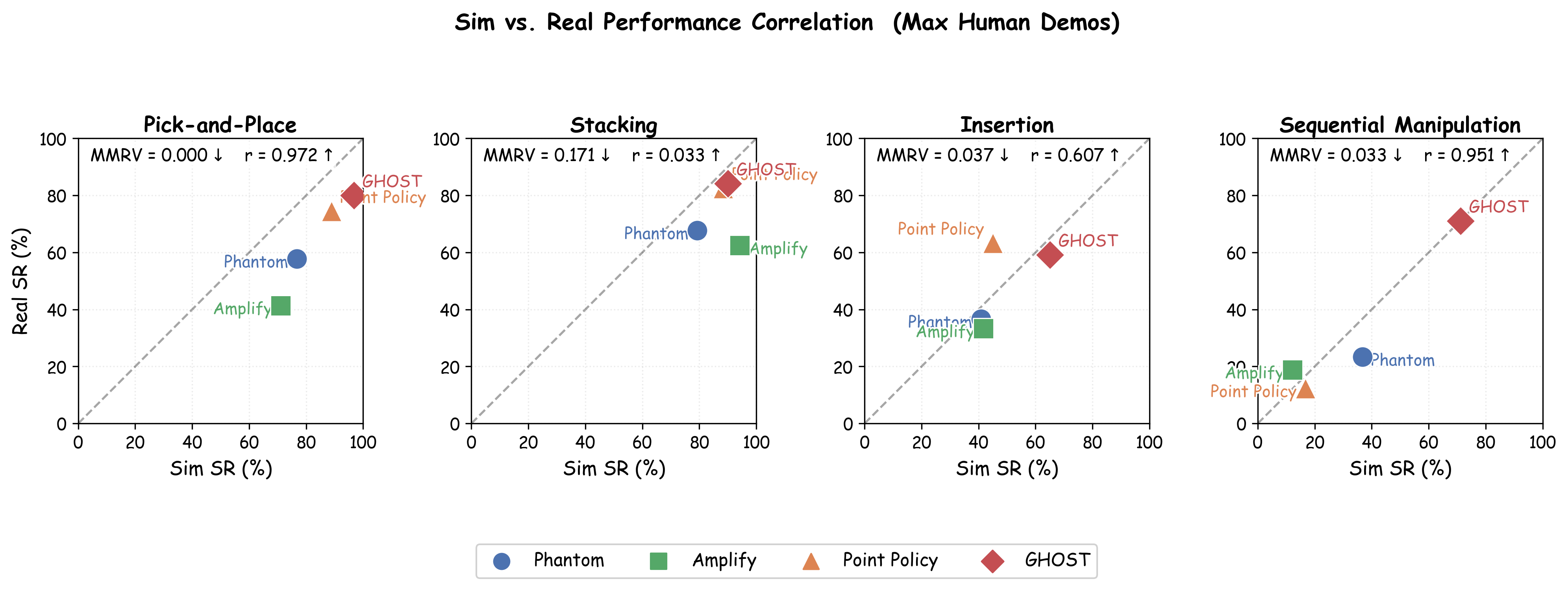}

    \caption{Per-task Sim-vs.-Real performance correlation across H2R methods.
Each panel shows the relationship between simulated and real-world
success rates for one task. The top row corresponds to robot-only
training (0 human demos), and the bottom row corresponds to training
with the maximum number of human demonstrations. Each point denotes
one method, with Pearson $r$ and MMRV reported per task.}
    \label{fig:per_task_mmrv}
\end{figure}

To better understand where Sim-vs.-Real discrepancies arise, we
further analyze performance correlation at the task level.

Table~\ref{tab:simreal_task_mmrv} reports per-task Mean Maximum Rank
Violation (MMRV), while Figure~\ref{fig:per_task_mmrv} visualizes the
corresponding Sim-vs.-Real success-rate relationship across methods.

We find that Sim-vs.-Real agreement is strongest for pick-and-place,
which achieves zero MMRV under both training settings. Sequential
manipulation and insertion also maintain relatively low ranking
violation, indicating that method ordering is largely preserved
between simulation and real-world evaluation.

The largest deviation occurs in the stacking task, particularly under
maximum human supervision. Compared with other tasks, stacking
requires more precise object alignment and stable contact interaction,
making success rates more sensitive to small differences in object
geometry, contact dynamics, and execution behavior between simulation
and the physical robot.

We additionally observe a larger Sim-vs.-Real performance shift for
AMPLIFY than for other methods. Because AMPLIFY predicts dense motion
tracks as an intermediate representation, its performance depends
strongly on the temporal smoothness and consistency of robot
trajectories. Compared with planner-generated trajectories in
simulation, real-world teleoperated demonstrations contain greater
motion variability, which can affect motion-track prediction quality
and lead to reduced downstream policy performance.

Importantly, these performance shifts do not always result in changes
in relative method ordering. In several cases, method rankings remain
stable despite moderate changes in absolute success rate. This
suggests that remaining Sim-vs.-Real gaps are often method-dependent
and primarily reflect differences in representation robustness to
real-world execution variation rather than limitations of the
benchmark itself.

\subsection{Ablation on Sim vs. Real Correlation}
To further characterize the source of remaining Sim-vs.-Real
discrepancy, we perform a sensitivity analysis by excluding
individual methods and tasks from the evaluation. Table~\ref{tab:simreal_ablation} summarizes Sim-vs.-Real metrics under
different subsets.

Excluding AMPLIFY consistently improves Pearson correlation,
Spearman correlation, and MMRV under both training settings, with the
largest improvement observed under maximum human supervision. This is
consistent with the per-task analysis and supports the observation
that AMPLIFY exhibits a larger Sim-vs.-Real performance shift than
other evaluated methods.

Excluding the stacking task also reduces MMRV, particularly in the
robot+human setting. This further confirms that stacking contributes
disproportionately to ranking inconsistencies across domains.

Overall, these ablation results indicate that observed Sim-vs.-Real
discrepancies are concentrated in a small number of method--task
interactions rather than uniformly distributed across H2RBench.
Despite these localized differences, the benchmark maintains strong
overall predictive value for comparative H2R evaluation before
real-world deployment.
\begin{table}[H]
\centering
\caption{
Sensitivity analysis of Sim-vs.-Real metrics under different subsets
of methods and tasks.
}
\label{tab:simreal_ablation}

\footnotesize
\setlength{\tabcolsep}{4pt}
\renewcommand{\arraystretch}{1.0}

\resizebox{\columnwidth}{!}{%
\begin{tabular}{llcccc}
\toprule
\textbf{Condition}
& \textbf{Subset}
& \textbf{N}
& \textbf{Pearson $r$ $\uparrow$}
& \textbf{Spearman $\rho$ $\uparrow$}
& \textbf{MMRV $\downarrow$} \\
\midrule

Robot Only
& Full Benchmark
& 16
& 0.971
& 0.940
& 0.014 \\

& Excluding AMPLIFY
& 12
& 0.984
& 0.970
& 0.003 \\

& Excluding Stack Bowls
& 12
& 0.971
& 0.986
& 0.005 \\

\midrule

Robot + Max Human
& Full Benchmark
& 16
& 0.894
& 0.851
& 0.060 \\

& Excluding AMPLIFY
& 12
& 0.924
& 0.902
& 0.007 \\

& Excluding Stack Bowls
& 12
& 0.887
& 0.890
& 0.023 \\

\bottomrule
\end{tabular}
}
\end{table}

\subsection{Metric Calculation}
\label{sec:metric_calculation}

To quantify Sim-vs.-Real correspondence, we report Pearson correlation,
Spearman rank correlation, and Mean Maximum Rank Violation (MMRV).

\paragraph{Pearson Correlation.}
Pearson correlation measures the linear relationship between simulated
and real-world success rates across method--task pairs:

\begin{equation}
r =
\frac{
\sum_{i=1}^{N}(x_i-\bar{x})(y_i-\bar{y})
}{
\sqrt{\sum_{i=1}^{N}(x_i-\bar{x})^2}
\sqrt{\sum_{i=1}^{N}(y_i-\bar{y})^2}
},
\end{equation}

where $x_i$ and $y_i$ denote the simulated and real-world success
rates for experiment $i$, and $\bar{x}$ and $\bar{y}$ are the
corresponding means.

\paragraph{Spearman Rank Correlation.}
Spearman correlation evaluates monotonic agreement between simulated
and real-world performance rankings:

\begin{equation}
\rho
=
1-\frac{6\sum_{i=1}^{N}d_i^2}{N(N^2-1)},
\end{equation}

where $d_i$ is the difference between the rank of experiment $i$ in
simulation and in the real world.

For both Pearson $r$ and Spearman $\rho$, we compute the correlation
over all $N=16$ method--task pairs within each evaluation condition
(robot-only or robot + max human demos). We use pooled computation
across tasks rather than per-task correlation because each individual
task contains only four methods, which leads to unstable rank-based
statistics with limited sample size. Aggregating across tasks provides
a more reliable estimate of overall Sim-vs.-Real correspondence while
capturing whether higher-performing policies in simulation also tend
to perform better in the real world.

\paragraph{Mean Maximum Rank Violation (MMRV).}
To quantify ranking inconsistency at the task level, we additionally
report Mean Maximum Rank Violation (MMRV), following
SIMPLER~\cite{li2024evaluating}:

\begin{equation}
\small
\mathrm{MMRV}
=
\frac{1}{T}
\sum_{t=1}^{T}
\frac{1}{N}
\sum_{i=1}^{N}
\max_{j \neq i}
\Bigg[
|r_i^{(t)} - r_j^{(t)}|
\cdot
\mathbf{1}
\Big(
(s_i^{(t)} < s_j^{(t)})
\neq
(r_i^{(t)} < r_j^{(t)})
\Big)
\Bigg]
\end{equation}

where $s_i^{(t)}$ and $r_i^{(t)}$ denote the simulated and real-world
success rates of method $i$ on task $t$, respectively, normalized to
$[0,1]$. $N$ denotes the number of methods, $T$ the number of tasks,
and $\mathbf{1}[\cdot]$ is the indicator function, which equals $1$
when the relative ordering between two methods differs between
simulation and real-world evaluation, and $0$ otherwise.

We compute MMRV independently for each task and report the average
across all $T$ tasks. Lower MMRV indicates stronger agreement in
relative method ordering between simulation and the real world.

\end{document}